\documentclass{article} 

\usepackage[compatibility=false]{caption}
\usepackage{subcaption} 

\usepackage[accepted]{mlsys2025}

\usepackage{amsmath,amsfonts,bm}

\def\eqref#1{equation~\ref{#1}}

\def\1{\bm{1}}

\DeclareMathAlphabet{\mathsfit}{\encodingdefault}{\sfdefault}{m}{sl}
\SetMathAlphabet{\mathsfit}{bold}{\encodingdefault}{\sfdefault}{bx}{n}

\usepackage{xcolor}
\usepackage[most]{tcolorbox}
\usepackage[framemethod=tikz]{mdframed}
\usepackage{xparse}
\usepackage{tikz}
\usepackage{varwidth}

\newtcbox{\bracketeq}{
  nobeforeafter, enhanced, colback=gray!10, frame empty, math upper,
  boxsep=2pt, left=8pt, right=8pt, top=4pt, bottom=4pt, arc=4pt,
  tcbox raise base, 
  overlay={
    \def\thk{1.0pt}   
    \def\stub{4pt}    
    \draw[line width=\thk] (frame.north west) -- (frame.south west);
    \draw[line width=\thk] (frame.north east) -- (frame.south east);
    \draw[line width=\thk] (frame.north west) -- ([xshift=\stub]frame.north west);
    \draw[line width=\thk] (frame.south west) -- ([xshift=\stub]frame.south west);
    \draw[line width=\thk] (frame.north east) -- ([xshift=-\stub]frame.north east);
    \draw[line width=\thk] (frame.south east) -- ([xshift=-\stub]frame.south east);
  }
}

\newtcbox{\boxeq}{%
  enhanced,
  on line,
  nobeforeafter,
  math upper,          
  frame empty,
  colback=gray!10,
  boxsep=2pt, left=4pt, right=4pt, top=4pt, bottom=4pt,
  arc=3pt,
  tcbox raise base,
  overlay={
    \def\r{3pt}         
    \def\thk{1.0pt}     
    \def\stub{4pt}      
    \def\wipe{1pt}      

    \draw[line width=\thk, rounded corners=\r]
      (frame.north west) rectangle (frame.south east);

    \fill[white]
      ([xshift=\stub,yshift=\wipe]frame.north west)
      rectangle
      ([xshift=-\stub,yshift=-\wipe]frame.north east);
    \fill[white]
      ([xshift=\stub,yshift=\wipe]frame.south west)
      rectangle
      ([xshift=-\stub,yshift=-\wipe]frame.south east);
  }
}

\tcbuselibrary{theorems,skins,breakable}

\tcbset{
  raise-lab/boxed theorem/.style={
    enhanced, breakable,
    colback=blue!5!white,
    colframe=blue!75!black,
    boxrule=1.5pt,
    arc=3mm,
    left=2mm, right=2mm,
    top=3mm,
    attach boxed title to top left={xshift=3mm, yshift=-3mm},
    boxed title style={
      colback=blue!25!white,
      colframe=blue!75!black,
      boxrule=1.5pt,
      left=1mm, right=1mm, top=0mm, bottom=0mm
    },
    fonttitle=\bfseries,
    coltitle=black,
    varwidth boxed title
  }
}

\makeatletter
\@ifundefined{c@theorem}{%
  \newtcbtheorem[number within=section]{boxdef}{Definition}{raise-lab/boxed theorem}{def}
  \newtcbtheorem[use counter from=boxdef]{boxassump}{Assumption}{raise-lab/boxed theorem}{assump}
  \newtcbtheorem[number within=section]{boxtheorem}{Theorem}{raise-lab/boxed theorem}{def}
}{%
  \newtcbtheorem[use counter from=theorem]{boxdef}{Definition}{raise-lab/boxed theorem}{def}
  \newtcbtheorem[use counter from=theorem]{boxassump}{Assumption}{raise-lab/boxed theorem}{assump}
  \newtcbtheorem[use counter from=theorem]{boxtheorem}{Theorem}{raise-lab/boxed theorem}{def}
}
\makeatother

\NewDocumentCommand{\boxdefc}{ O{} +m }{%
  \begin{boxdef}{#1}
    #2
  \end{boxdef}%
}

\NewDocumentCommand{\boxtheoremc}{ O{} +m }{%
  \begin{boxtheorem}{#1}
    #2
  \end{boxtheorem}%
}

\makeatletter
\@ifpackageloaded{cleveref}{%
  \crefname{def}{Definition}{Definitions}
  \Crefname{def}{Definition}{Definitions}
  \crefname{assump}{Assumption}{Assumptions}
  \Crefname{assump}{Assumption}{Assumptions}
}{}
\makeatother

\usepackage{hyperref}
\usepackage{url}
\usepackage{amssymb} 
\usepackage{amsthm}
\usepackage{enumitem}
\usepackage{comment}
\usepackage{graphicx}
\usepackage{booktabs,multirow}

\usepackage{fancyvrb}
\usepackage{verbatim}
\usepackage{listings}
\usepackage{listings}
\usepackage{xcolor}

\usepackage{cleveref}
\usepackage{etoolbox}
\AtBeginEnvironment{corollary}{\upshape}
\usepackage{xfrac}
\usepackage[table]{xcolor}
\newcommand{\shad}{\cellcolor{gray!25}} 
\newcommand{\streak}{\tfrac{\text{Tokens}}{\text{Draft}}}

\newcommand{\Qdiff}{Q^{\mathrm{diff}}}

\newcommand{\jacob}[1]{\textcolor{blue}{[Jacob: #1]}}

\theoremstyle{plain}
\newtheorem{theorem}{Theorem}[section]

\theoremstyle{definition}
\newtheorem{definition}[theorem]{Definition}

\theoremstyle{remark}

\mlsystitlerunning{Scaling Diffusion Drafter Alignment for Faster Speculative Decoding}

\usepackage[ruled,vlined]{algorithm2e}

\makeatletter
\@ifundefined{tcp}{}{
  \let\old@tcp\tcp
  \renewcommand{\tcp}[2][]{\old@tcp*[r]{\textnormal{#2}}}
}
\SetCommentSty{textnormal}
\makeatother

\begin{document}

\twocolumn[
\mlsystitle{SpecDiff-2: Scaling Diffusion Drafter Alignment For Faster Speculative Decoding}

\mlsyssetsymbol{equal}{*}

\begin{mlsysauthorlist}
\mlsysauthor{Jameson Sandler}{equal,uva}
\mlsysauthor{Jacob K.~Christopher}{equal,uva}
\mlsysauthor{Thomas Hartvigsen}{uva}
\mlsysauthor{Ferdinando Fioretto}{uva}
\end{mlsysauthorlist}

\mlsysaffiliation{uva}{Department of Computer Science, University of Virginia, Charlottesville, USA}

\mlsyscorrespondingauthor{Nando Fioretto}{fioretto@virginia.edu}
\mlsyscorrespondingauthor{Jameson Sandler}{jmz4ds@virginia.edu}
\mlsyscorrespondingauthor{Jacob K.~Christopher}{csk4sr@virginia.edu}

\mlsyskeywords{Machine Learning, MLSys}

\vskip 0.3in

\begin{abstract}
    Speculative decoding has become the standard approach for accelerating Large Language Model (LLM) inference. It exploits a lossless draft-then-verify procedure to circumvent the latency of autoregressive decoding, achieving impressive speed-ups.
    Yet, current speculative decoding approaches remain limited by two fundamental bottlenecks: \textbf{(1)} the autoregressive dependency during drafting which limits parallelism, and \textbf{(2)} frequent rejections of draft tokens caused by misalignment between the draft and verify models. 
    This paper proposes \emph{{SpecDiff-2}}, a novel framework to jointly address these two bottlenecks. It leverages discrete diffusion as a non-autoregressive drafter to address bottleneck (1) and develops novel techniques to calibrate discrete diffusion drafters with autoregressive verifiers, addressing bottleneck (2). Experimental results across a comprehensive benchmark suite show that \emph{SpecDiff-2} achieves a new state-of-the-art across reasoning, coding, and mathematical benchmarks, improving tokens-per-second by up to an average of $+55\%$ over previous baselines and obtaining up to $5.5\times$ average speed-up over standard decoding, without any loss of accuracy. 
\end{abstract}
]

\printAffiliationsAndNotice
\mlsysEqualContribution

\newlength{\panelH}
\setlength{\panelH}{2.30in}

\section{Introduction}

The performance of large language models (LLMs) has rapidly improved alongside increases in both model size and computational budget.
Inference-time compute scaling, for instance, has been shown to significantly improve performance on complex tasks, with techniques such as long chain-of-thought reasoning and self-consistency, increasing success rates by allocating more computation at inference time \citep{wei2022chain, wang2022self, zelikman2022star, schick2023toolformer}. 
However, these gains are obtained at the cost of higher wall-time latency; because LLMs, being predominantly based on autoregressive (AR) architectures, produce tokens sequentially, deeper reasoning chains will translate into slower response times. As a result, the depth of problem solving that can be realized, in practice, is constrained by wall-time budgets.

\begin{figure}[t]
  \centering
  \includegraphics[width=0.9\linewidth]{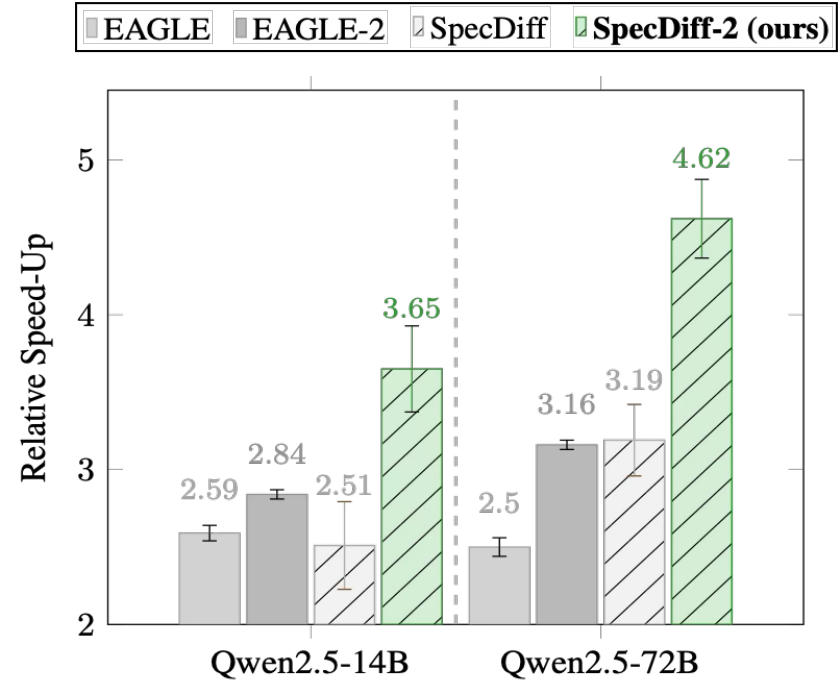}
\vspace{-8pt}
\caption{Throughput increase (y-axis) relative to vanilla inference across SoTA acceleration algorithms. Tested on Math500 across 14B and 72B, Qwen2.5-Instruct models. The figure also shows how 
the \emph{aligned} drafter in \emph{SpecDiff-2} outperforms the base \emph{SpecDiff} drafter by over $40\%$.}
\label{fig:motivation}
\vspace{-15pt}
\end{figure}

Consequently, inference-time acceleration techniques have become an important area of research. The goal is to reduce the latency of LLM generation without sacrificing model quality. In particular, \emph{speculative decoding} has emerged as a leading framework for overcoming this constraint \cite{leviathan2023fast, chen2023accelerating}. This framework is built around a draft-then-verify procedure: a small drafter model proposes multiple tokens, and a heavier verifier model evaluates the proposal in parallel to accept the longest matching sequence. 
When the draft model can generate ``high-quality'' tokens quickly, multiple tokens can be accepted per verification cycle, recovering the outputs of vanilla decoding while reducing end-to-end latency. 
However, as also noted by \citealt{yan2024decoding}, the realized speed-up depends on two key factors: \textbf{(1)} \emph{the drafter latency}, since generating draft proposals incurs extra time and the drafter must be fast relative to the verifier; and \textbf{(2)} \emph{the drafter-verifier alignment}, since misaligned proposals are likely to be rejected, forcing regeneration from the rejection point onward. 
Thus, the realized throughput hinges on increasing {the expected acceptance per cycle} while keeping {drafter cost low} relative to the verifier.

To address the aforementioned challenges, this paper develops \emph{SpecDiff-2}, a speculative decoding system that targets both bottlenecks simultaneously. The approach leverages diffusion language models (DLMs)
as non-autoregressive drafters to address bottleneck (1) and develops novel alignment mechanisms to calibrate diffusion drafters with verifiers at train and test-time, addressing bottleneck (2). 

Discrete diffusion models generate text by iteratively transitioning the token space toward a fluent sequence in a small, fixed number of steps \cite{sahoo2024simple,shi2024simplified}. Each denoising step updates all token positions in parallel, so the drafting cost depends primarily on the number of steps rather than the sequence length. This makes discrete diffusion particularly suitable for addressing bottleneck (1): it eliminates the token-by-token dependency of autoregressive drafting, exploits accelerator-friendly batching, and delivers low latency for proposing multi-token drafts \cite{christopher2025speculative}.
However, \emph{diffusion drafters and autoregressive verifiers produce fundamentally different objects}. Diffusion models learn a joint distribution over entire sequences through denoising trajectories, whereas autoregressive models learn local next-token conditionals tied to causal prefixes. As a result, raw diffusion samples can be well-formed globally yet miscalibrated locally with respect to the verifier token-wise decisions. {\em Aligning these two views is nontrivial and requires mechanisms that bridge joint-generation behavior with prefix-conditional acceptance.}
SpecDiff-2 achieves this goal via two complementary mechanisms: a train-time (de-facto, fine-tuning) procedure,Ju called \emph{streak-distillation}, that improves proposer alignment with the verifier by targeting theoretical acceleration, and a test-time acceptance mechanism, called \emph{self-selection acceptance}, that uses the verifier to select drafts that maximize throughput. The consequent framework results in significant throughput improvements ($>42\%$) over prior diffusion-based drafters, as illustrated in Figure \ref{fig:motivation},
and over 300\% improvements over vanilla decoding, without degrading accuracy.

\textbf{Contributions.}
This work makes the following key contributions:
{\bf (1)} It introduces a parallel drafting mechanism based on discrete diffusion models for speculative decoding, {\bf (2)} to cope with the misalignment between diffusion drafters and autoregressive verifiers, it develops \emph{streak-distillation}, a novel train-time alignment method that encourages the drafter to produce long streaks of accepted tokens, {\bf (3)} this procedure is coupled with \emph{self-selection acceptance}, a test-time mechanism that selects drafts most consistent with the verifier. {\bf (4)} Finally, it demonstrates state-of-the-art throughput across a comprehensive benchmark suite, improving tokens-per-second by up to an average of $+55\%$ over previous baselines and obtaining up to $5.5\times$ average speed-up over standard decoding, without any loss of accuracy. 

\section{Prelim: Speculative Decoding}
\label{sec:prelim}

Speculative decoding (SD) algorithms are built on two language models, a small \emph{drafter} model $Q$, and a heavier target \emph{verifier} $P$ \cite{leviathan2023fast, chen2023accelerating}. 
These models generate a sequence of tokens $\bm{x}_{1:n} = (x_1, \ldots, x_n)$, denoted $\bm{x}$ when clear from context, within a vocabulary $\mathcal V$. For a prefix $\bm{s}\in\mathcal V^\star$, the models expose next-token posteriors $q(\cdot\mid \bm{s})$ and $p(\cdot\mid \bm{s})$, associated, respectively, to models $Q$ and $P$.
Generation proceeds in blocks of size $\gamma$: the drafter proposes $\bm{x}_{1:\gamma}\sim Q(\cdot \mid \bm{s})$ while the verifier scores all positions in parallel. 
Tokens are committed \emph{left-to-right} using the standard lossless acceptance rule \cite{leviathan2023fast}: 

\textit{For a drafted token $x_i$ at a given position $i$,}
\begin{equation}
  \label{eq:sd_accept}
  \Pr(\text{accept } x_i) = \min \left\{ 1, \frac{p(x_i \mid \bm{s})}{q(x_i \mid \bm{s})} \right\},
\end{equation}
\textit{where $\bm{s}$ here, and throughout the paper, denotes the prefix preceding $x_i$: i.e., $\bm{s} = \bm{x}_{1:i-1}$ (for $i>1$). }

This rule results in immediate acceptance whenever $q(x_i \mid \bm{s}) \leq p(x_i \mid \bm{s})$. 
On the first rejection at position $k$, the remaining draft is discarded and a \emph{replacement} is drawn from the normalized residual:
\[
  r(x)\ \propto\ \bigl[p(x \mid \bm{s}) - q(x \mid \bm{s})\bigr]_+.
\]
These rules are distributionally faithful: the final transcript matches vanilla decoding from $P$ while enabling parallel proposal and verification \cite{leviathan2023fast}, (see Appendix \ref{A:sd-invariance} for proof).

\section{Acceptance \& Distillation}
\label{sec:motivation}

The acceptance rate governs the realized speed-up in draft-then-verify decoding. 
\begin{definition}[SD Acceptance Rate]
  \label{def:acceptance-rate}
  For a prefix $\bm{s}$ and drafter and verifier pairs $Q$ and $P$, the speculative-decoding \emph{acceptance rate}, $\alpha(\bm s)$, over all prefixes $\bm{s}=\bm{x}_{1:i-1}$, is defined as:
  \begin{align}
    \label{eq:acceptance-rate}
  \alpha(\bm{s}) &\triangleq\ 
     \Bigl[1 - 
    \underbrace{\tfrac{1}{2}\sum_{x_i \in \mathcal V} \mid P(x_i \mid \bm{s}) - Q(x_i \mid \bm{s})\mid }_{\textrm{TV}(P,Q)}\Bigr].
  \end{align}
\end{definition}

\begin{figure}[t]
  \begin{subfigure}[t]{0.85\linewidth}
    \centering
    \includegraphics[height=2.2in]{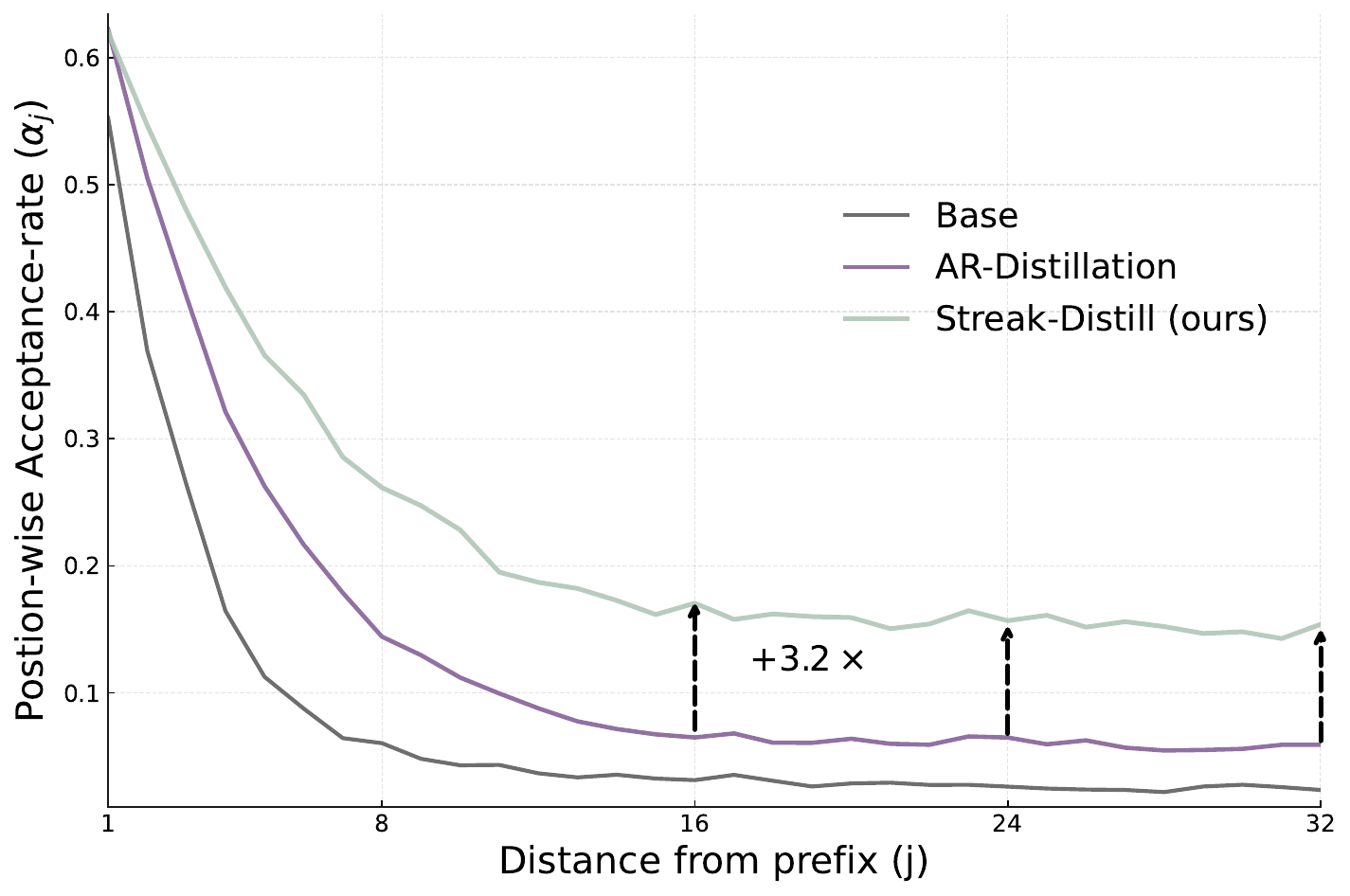}
  \end{subfigure}
  \caption{Position-wise acceptance (y-axis, higher is better) against prefix index j (x-axis, distance from prefix $\bm{s}$). Autoregressive alignment (AR-Distillation above) struggles to align later window indices to verifier (Qwen2.5-14B-Instruct), given lack of window-wise consideration. Both algorithms ran for 20k gradient steps, tested on the Alpaca benchmark.}
  \label{fig:sd_vs_ar}
  \vspace{-20pt}
\end{figure}

Further, the acceptance-rate exactly controls the throughput that speculative-decoding benefits from. Let $\alpha_j(\bm{s})$ denote the position-wise acceptance, that is, the probability that the $j$-th drafted token is accepted \emph{conditional} on the first $j\!-\!1$ drafted tokens having been accepted. Under the standard product-of-accepts identity, the expected number of committed tokens from a draft of length $\gamma$ is:
\begin{equation}
  \label{eq:expected-streak}
    \tfrac{\text{Tokens}}{\text{Draft}}(\gamma, \bm{s}) = \mathbb{E}_{\bm{x}_{1:\gamma}\sim Q(\cdot\mid\bm{s})}\bigl[\sum_{m=1}^{\gamma}\prod_{j=1}^{m} \alpha_j ({\bm{s}} \circ \bm{x}_{1:{j-1}})\bigr],
\end{equation}
where $\circ$ denotes sequence concatenation.

Thus, improvements in the $\alpha_j(\bm{s})$ translate monotonically into longer accepted streaks and higher throughput at fixed drafter cost.

The acceptance-rate $\alpha$ is then a measure of ``alignment'', scaling inversely to the total-variation distance (TV) between $P$ and $Q$. \emph{Increasing the alignment} between $P$ and $Q$, thus translates to \emph{increasing inference throughput}.
Therefore, distillation based alignment methods that target TV reduction directly have become a popular practice for accelerating speculative decoding in autoregressive drafters \cite{zhou2023distillspec}.
However, these methods commonly treat the per-position acceptances as exchangeable and only optimize the alignment immediately following the prefix (i.e., $\max[\alpha_1({\bm{s}} \circ \bm{x}_{1})]$). 
Indeed, when the loss function is defined over a windows of $\textrm{size} = 1$, this ``simplified'' minimization produces an identical minimizer to the true acceptance criteria (derivation reported in Appendix \ref{A:pos_alpha_identity}). 

Yet, the position-wise alignment levels vary significantly for diffusion drafters, and thus, the full window of draft predictions must be carefully optimized to obtain the expected speed-up. 
This is illustrated empirically in Figure \ref{fig:sd_vs_ar}, showing the position-wise acceptance $\alpha_j$ (y-axis) against the draft index $j$ (x-axis) for three settings: the base diffusion drafter (``Base''), the diffusion drafter aligned via an autoregressive objective (``AR-distillation''), and the proposed streak-aware objective (``Streak-distillation'', Section~\ref{sec:streak-distillation}).
Notice that using an AR-style alignment concentrates gains at early positions and degrades rapidly with $j$, leaving later tokens poorly calibrated. This motivates the need for alignment procedures that align over the full window. Our approach contrasts this behavior by correctly aligning the later indices in the draft window at an average of $3.2\times$ greater acceptance than the AR approach at these positions. 

\section{Speculative Diffusion Decoding}
\label{sec:specdiff}

Before introducing our novel alignment tecniques, this section reviews Speculative Diffusion Decoding (or SpecDiff) \cite{christopher2025speculative}. SpecDiff builds on the speculative decoding paradigm by substituting the autoregressive drafter model, $Q$, for a low-latency \emph{diffusion drafter}. 
This exploits a key property of parallel drafting: proposed tokens in a window of size $\gamma$ are generated \emph{simultaneously}, so drafter cost depends primarily on the number of denoising steps of the diffusion model rather than on $\gamma$, addressing the latency bottleneck noted earlier.
Below the paper writes $Q$ and $\Qdiff$ to distinguish between autoregressive and diffusion drafters, respectively. 
In the following, we briefly review discrete diffusion models for text generation, and, in particular, masked discrete diffusion models on account of their high performance on language-modeling tasks \citep{sahoo2024simple, shi2024simplified}.

\paragraph{Masked-discrete diffusion language models.} 
Discrete diffusion language models implement non-autoregressive drafting via corruption-denoising steps. A forward process produces a partially corrupted sequence $\bm{x}^{(t)}$ for $t\in[0,1]$ from a clean sequence $\bm{x}$, and a denoiser $\Qdiff$ predicts token distributions conditioned on $\bm{x}^{(t)}$. 
In masked diffusion models (MDMs) \cite{sahoo2024simple}, a special $[\mathrm{MASK}]$ token is introduced and each position $i$ is independently replaced by $[\mathrm{MASK}]$ with probability $1-\varepsilon(t)$; otherwise it remains $x_i$. The denoiser outputs $\Qdiff(\cdot\mid \bm{x}^{(t)}; \theta)_i$ for all positions in parallel and is trained with masked cross-entropy over the currently masked set $M(t)$:
\begin{equation*}
\mathcal L(\theta) = 
  \mathbb E_{\bm{x}^{(t)}}\!\left[\frac{1}{|M(t)|}\sum_{i\in M(t)}
    - \log \Qdiff \big(x_i \mid \bm{x}^{(t)}; \theta \big)_i \right].
\end{equation*}
Drafting at prefix $\bm{s}$ with window $\gamma$ then proceeds as follows. It first populates the draft tokens $\bm{x}_{1:\gamma}$ with $[\mathrm{MASK}]$ tokens; a small number of denoising steps (often a single step in practice) then is applied to yield a joint proposal $\bm{x}_{1:\gamma} \sim \Qdiff(\cdot \mid \bm{s} \circ [\mathrm{MASK}]^{1:\gamma})$, produced in parallel across positions.
The paper denotes $\Qdiff(\cdot | \bm{s})$ to express this joint process, with parameters $\theta$ generally omitted where irrelevant, though this should not be confused with an autoregressive conditional. 

As motivated in the previous section, while speculative diffusion delivers low and predictable drafter latency, \emph{it does not by itself solve alignment}. Diffusion drafters are trained to model joint denoising distributions over blocks, whereas the verifier evaluates prefix-conditional next-token posteriors; miscalibration at the token level therefore reduces acceptance even when joint samples appear fluent. The next sections introduce alignment objectives and test-time mechanisms tailored to MDM drafters (Section~\ref{sec:alignment}) and report the resulting throughput gains (Section~\ref{sec:experiments} and \ref{sec:results}).

\section{Aligning Diffusion Drafters For Accelerated Inference}
\label{sec:alignment}

We are now ready to introduce the key contribution of this work: two alignment mechanisms tailored to diffusion drafters, a \emph{train-time} (finetuning) objective that scales with distillation compute (Section \ref{sec:streak-distillation}) and a \emph{test-time} selection rule that scales with verifier compute (Section \ref{subsec:test-time}). Importantly, they both act solely on 
$\Qdiff$, while the verifier $P$ remain frozen throughout. 
Crucially, these mechanisms yield robust drafter–verifier alignment that \emph{generalizes beyond the finetuning data}: indeed, all evaluations are conducted on datasets disjoint from those used in finetuning (see Section~\ref{sec:results}).
Together, and coupled with the use of speculative diffusion, they give rise to SpecDiff-2, a new state of the art speculative decoder.

\begin{figure}[t]
    \centering
    \includegraphics[width=0.99\linewidth]{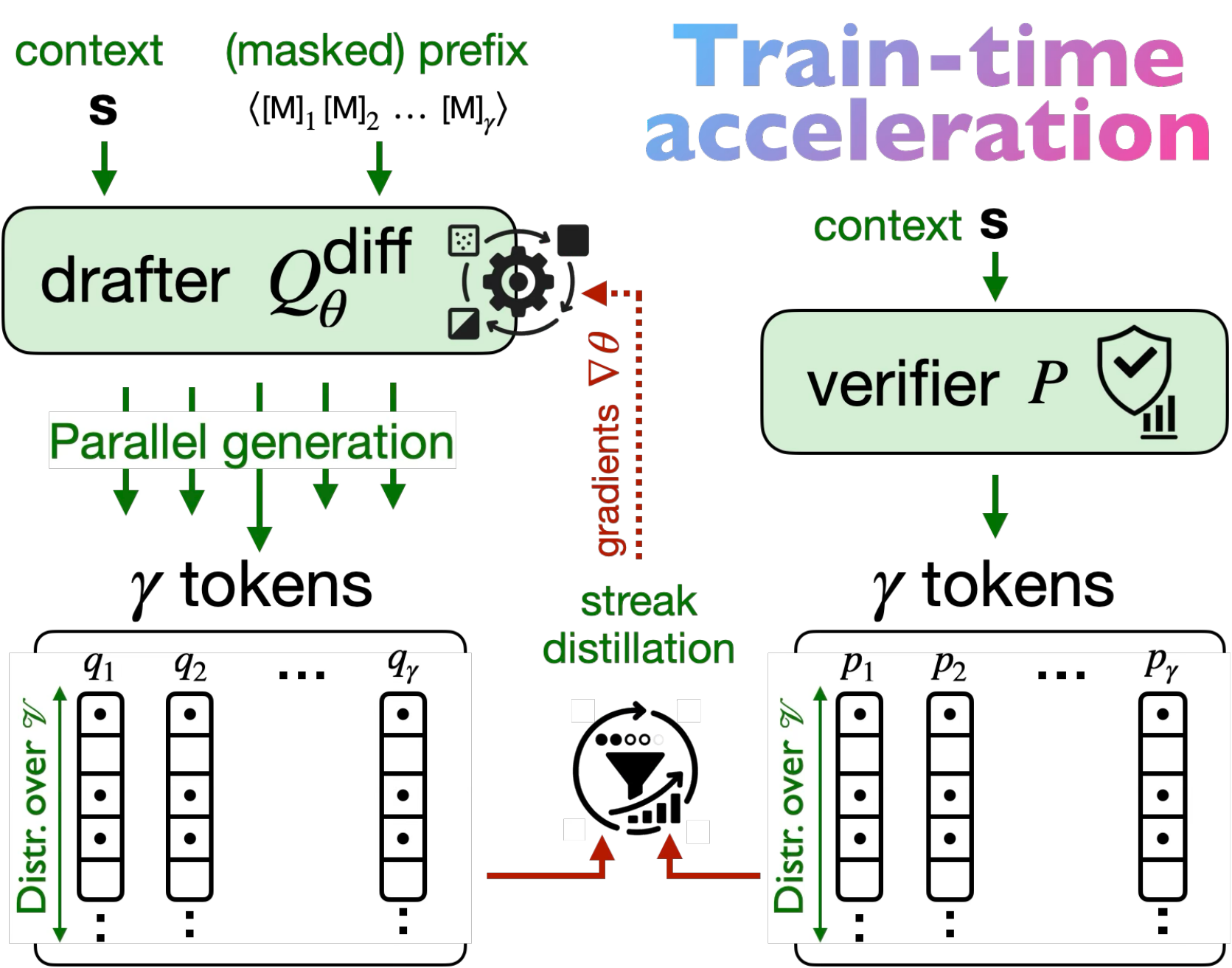}
    \caption{
    {\bf Train-time acceleration} of \emph{SpecDiff-2}. 
    $\Qdiff_\theta$ (parameters $\theta$) via the streak-distillation equation (see Definitzion \ref{def:sd-objective}). References are supplied via $P$, then logits $q_1, \ldots, q_\gamma$ are computed with $\Qdiff$. Streak-distill equation is computed, and $Q$ is updated via gradient ascent over expected throughput.
    }
    \label{fig:streak_distillation}
    \vspace{-10pt}
\end{figure}

\subsection{Train-time alignment: Streak-distillation}
\label{sec:streak-distillation}

The goal of this section is to align diffusion drafters to the verifier in a manner that \emph{directly} improves the expected accepted streak in 
$\tfrac{\text{Tokens}}{\text{Draft}}(\gamma, \bm{s})$, defined in 
\Cref{eq:expected-streak}, while keeping training tractable. 

The construction begins with a proxy for acceptance, \emph{greedy acceptance}, that yields a smooth training signal. At a prefix $\bm{s}$, the probability of accepting a drafted token is set to the verifier probability:
$$ \Pr(\text{accept } x_i \mid \bm{s})=P( x_i \mid \bm{s}).$$ 
Note that this scheme does not use the drafter posterior during verification and acts only as an analytical device for deriving a distillation objective. 
Under this greedy scheme, the position-wise acceptance $\tilde{\alpha}_j$ along a teacher path $\bm{x}_{1:j-1}\!\sim\! P(\cdot\mid \bm{s})$ reduces to the product:
\begin{equation}
  \label{eq:alpha-greedy}
  \tilde{\alpha}_j(\bm{s})
  \!=\!
  \operatorname{\mathbb{E}}
  \!\left[
  \sum_{x_j \in \mathcal V} 
  P(x_j \!\mid\! \bm{s} \!\circ\! \bm{x}_{1:j-1}) \, 
  \Qdiff(x_j \!\mid\! \bm{s}) 
  \right],
\end{equation}

where the expectation is taken over $\bm{x}_{1:j-1}$. Note that the diffusion drafter is not conditioned on $\bm{x}_{1:j-1}$, as all tokens are generated in parallel. 
Since the inner sum in \Cref{eq:alpha-greedy} is a dot product between two categorical distributions over $\mathcal V$, it can be rewritten as an expectation w.r.t.~either distribution. Thus, the dependency from the verifier $P$ can be made implicit by absorbing it into the sampling distribution for $x_j$, yielding:
\begin{subequations}
\label{eq:streak-equiv} 
\begin{align}
  \tilde{\alpha}_j(\bm{s})
  \!&=\!
  \operatorname{\mathbb{E}}_{\bm{x}_{1:j-1}}
  \left[
  \operatorname{\mathbb{E}}_{
  {x_j \sim  P(\cdot\mid \bm{s} \circ \bm{x}_{1:j-1})}}
  \!\left[
      Q^{\text{Diff}}(x_j\mid \bm{s})
  \right] \right] \label{eq:streak-equiv-a} \\
 &= \!
  \operatorname{\mathbb{E}}_{\bm{x}_{1:j-1}}
  \left[
  \operatorname{\mathbb{E}}_{{x_j \sim  Q^{\text{diff}}(\cdot\mid \bm{s})}}
  \!\left[
      P(x_j \!\mid\! \bm{s} \circ \bm{x}_{1:j-1})
  \right]
  \right]. \!\!
  \label{eq:streak-equiv-b} 
\end{align}
\end{subequations}
The outer expectation remains over teacher prefixes $\bm{x}_{1:j-1}$ drawn from the verifier chain, while the inner expectation may be evaluated by sampling $x_j$ either from the verifier $P$ (i.e., during distillation, see Definition \ref{def:sd-objective}) or from the drafter $\Qdiff$ (i.e., during verification, see Section \ref{subsec:test-time}).

\noindent\textit{This identity is important}, as it permits a pathwise re-expression of the streak proxy and leads directly to a tractable training objective. Substituting \Cref{eq:streak-equiv-a} 
into the product-of-accepts construction of \Cref{eq:expected-streak} yields a pathwise estimator for 
$\tfrac{\text{Tokens}}{\text{Draft}}(\gamma, \bm{s})$ 
in which the inner product is evaluated along teacher trajectories. Maximizing this quantity across prefixes and verifier continuations yields the \emph{streak-distillation} objective.

\begin{definition}[Streak-distillation]
\label{def:sd-objective}
\textit{Let $P$ be a frozen verifier and $\Qdiff_\theta$ a diffusion drafter with position-wise marginals $q_j(\cdot \mid \bm{s}; \theta)$. The streak-distillation objective is}
{\small
\begin{equation}
 \!\!\!\tfrac{\text{Tokens}}{\text{Draft}}\!\big(\gamma, \bm{s}\big)
    \triangleq
\mathbb E_{\bm{s}}\,
\mathbb E_{\bm{x}_{1:\gamma}\sim P(\cdot \mid  \bm{s})}\!
\left[
\sum_{m=1}^{\gamma}\ \prod_{j=1}^{m} q_j\!\big(x_j\mid \bm{s}; \theta\big)
\right].\!\!\!
\label{eq:sd-objective}
\end{equation}
}
\end{definition}

This \emph{streak-distillation} objective mirrors the ideal target in \Cref{eq:expected-streak}: the product-of-accepts is preserved, and each factor is replaced by the greedy acceptance term evaluated at verifier tokens. Thus, \emph{the goal is to optimize over all the draft window}, and this is done via the tractable surrogate in \Cref{def:sd-objective} rather than the intractable path-dependent accept/reject process. Figure \ref{fig:streak_distillation} illustrates this procedure.

The same analysis also suggests a complementary lever at inference: selecting among multiple candidate drafts using the streak-oriented proxy can further enlarge the accepted prefix under a fixed verifier, as discussed next.

\begin{figure}[t]
    \centering
    \includegraphics[width=0.99\linewidth]{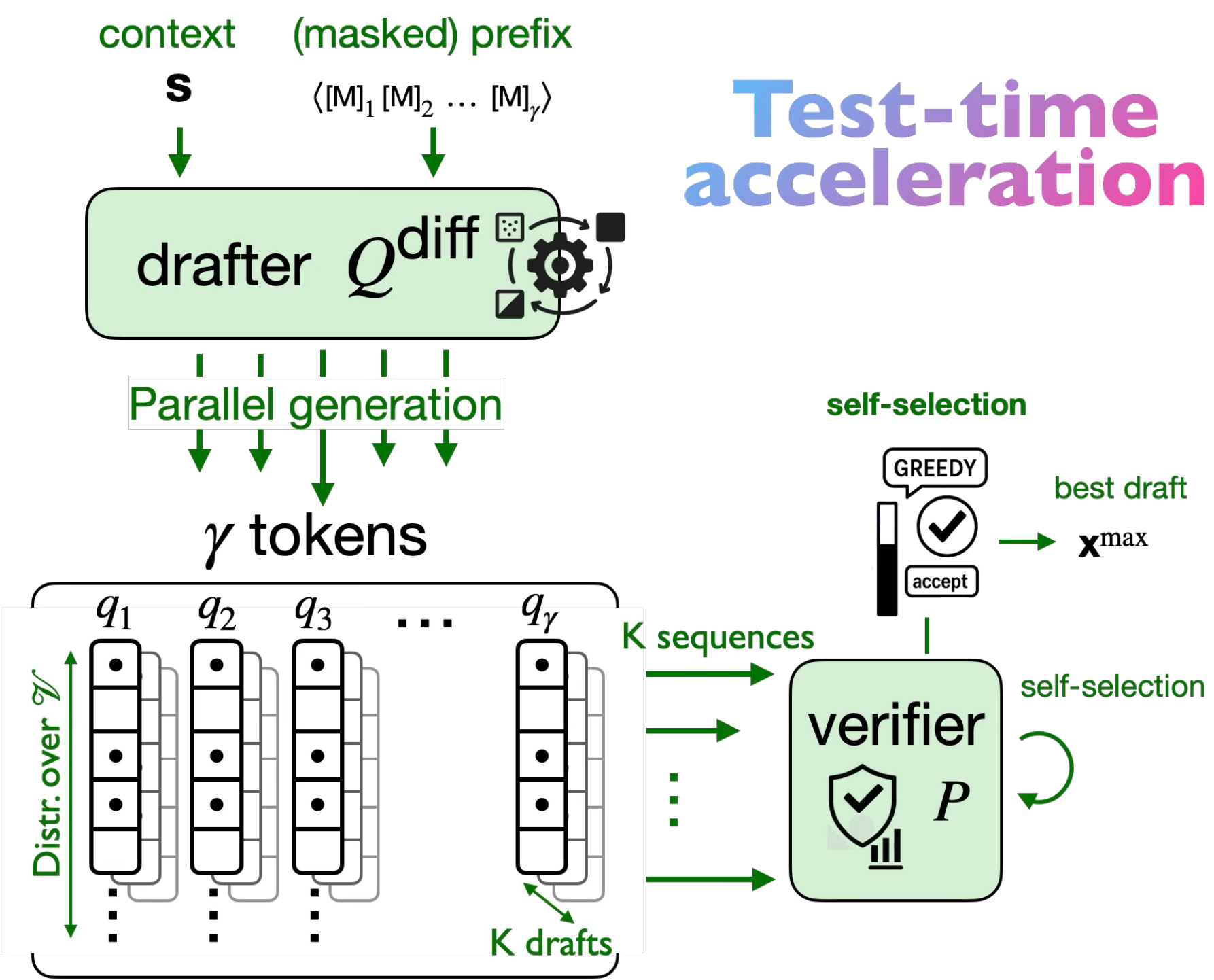}
    \caption{
    {\bf Test-time acceleration} of \emph{SpecDiff-2}. 
    For prefix $\bm{s}$ and draft of size $\gamma$, the drafter generates marginals $q_1, \ldots, q_\gamma$, and expands them into $K$ discrete drafts of length $\gamma$ via sampling. Then $P$ is used to select $\bm{x}^{\max}$, the best draft (\emph{self-selection}) to produce the final sequence.
    }
    \label{fig:test-time-graphic}
    \vspace{-15pt}
\end{figure}

\subsection{Test-time alignment: Self-selection acceptance}
\label{subsec:test-time}

While streak-distillation improves drafter-verifier alignment at train-time, additional gains can be obtained by leveraging verifier compute at test-time. 
Prior speculative methods (e.g., EAGLE-2 and beyond) realize this with autoregressive multi-path expansion, generating several proposal paths with $Q^{\text{ar}}$ to improve the likelihood the verifier will accept a continuation \citep{li2024eagle2}. 
However, this test-time approach has some drawbacks when using autoregressive drafters. 
Firstly, the drafting complexity increases: if $N$ tokens are drafted across $K$ distinct paths, then under ideal branching $N \propto \log (K \cdot \gamma)$ and drafting compute scales linearly with the number of tokens ${\cal{O}}(N)$ (for fixed $\gamma$). 
Further, generation of multiple paths must happen in a parallel tree-like manner, increasing complexity.

In contrast, this paper argues that diffusion drafters admit a considerably more simple and efficient multi-draft regime. A single denoising pass exposes all position-wise marginals, from which multiple joint drafts can be sampled with negligible additional cost. This observation motivates a \emph{test-time mechanism} that uses the verifier to select among these drafts before lossless verification.

\SetAlgoNlRelativeSize{-1}
\SetCommentSty{textnormal}
\SetNlSty{textbf}{}{}  

\begin{algorithm}[t]
\caption{Self-selection acceptance}\label{alg:self-select}
\SetKwInOut{Input}{Input}
\SetKwInOut{Output}{Output}
\Input{prefix $\bm{s}$; verifier $P(\cdot \!\mid\! \bm{s},\cdot)$; drafter $\Qdiff(\cdot \!\mid\! \bm{s})$; integers $K,\gamma$}
\Output{Final generation $\bm{y} \in \mathcal{V}^{\leq \gamma}$ from $P(\cdot \!\mid\! \bm{s})$}
$(q_1,\ldots,q_\gamma) \sim \Qdiff(\cdot \mid \bm{s})$\emph{ (draft marginals)}\\
\For{$k \gets 1$ \KwTo $K$ \emph{ (in parallel)}}{
  $\bm{x}_{1:\gamma}^{\,k} \sim (q_1,\ldots,q_\gamma)$\emph{ (select best draft)}\\
  $\tau_k \gets \frac{\mathrm{Tokens}}{\mathrm{Draft}}(\bm{x}_{1:\gamma}^{\,k}, \bm{s})$\emph{ (compute throughput)}\\
}
$(\bm{x}^{\max}, \bm{y}) \sim (\arg\max_k \tau_k, [\,])$\emph{ (select best draft)}\\
$\bm{y} \sim [\ ]$\;
\For{$i \sim 1$ \KwTo $\gamma$}{
  $p_i \gets P(\bm{x}^{\max}_{i} \mid \bm{s}\circ \bm{y})$\;
  $b \sim \mathrm{Bernoulli}(p_i)$\;
  \eIf{$b = 1$}{
    $\bm{y} \gets \bm{y}\circ \bm{x}^{\max}_{i}$\emph{ (Accept drafted token)}\\
  }{
    $x_i \sim \dfrac{P(\cdot\mid \bm{s}\circ\bm{y})}{1 - P(\bm{x}^{\max}_{i}\mid \bm{s}\circ\bm{y})}$\emph{ (Replace token)}\\
    $\bm{y} \gets \bm{y}\circ x_i$\;
  }
}
\Return{$\bm{y}$}
\end{algorithm}

\textbf{Self-selection acceptance.}
The proposed test-time procedure, called  \emph{self-selection acceptance}, benefits from the following properties: {\bf (1)} it requires minimal adaptation from vanilla Speculative Diffusion; {\bf (2)} the diffusion drafters provide both time and compute efficient drafting, where for $K$ paths, drafting compute scales with ${\cal O}(1)$ (for fixed $\gamma$); and {\bf (3)} the proposed approach scales with $K$ (up to the model drafting size ability), allowing additional  throughput with negligible sequential overhead in $K$, as illustrated in Figure \ref{fig:ssa-scaling} (discussed later in Section \ref{subsec:ablations}).

First, note that diffusion models bring a unique \emph{multi-draft capability} for efficient drafting of several paths.
In the speculative diffusion setting (see Section~\ref{sec:specdiff}), a single denoising step of $\Qdiff$ over $[\mathrm{MASK}]^\gamma$ produces position-wise marginals $\{q_j(\cdot\mid \bm{s})\}_{j=1}^{\gamma}$. A joint draft is then obtained by independent sampling,
\(
\bm{x}_{1:\gamma}\ \sim\ \prod_{j=1}^{\gamma} q_j(\cdot\mid \bm{s}),
\)
at negligible cost relative to the neural forward pass.\footnote{Sampling is inexpensive relative to the denoising computation and parallel over $j$.}
Repeating this sampling $K$ times yields $\{\bm{x}^{(k)}_{1:\gamma}\}_{k=1}^{K}$ once again with negligible sequential overhead in $K$. The proposed self-selection method ranks these candidates using a streak-oriented verifier score and selects the best draft for lossless verification. The approach is illustrated in Figure~\ref{fig:test-time-graphic}.

To evaluate which of the $K$ generated drafts exhibits the highest alignment with the distribution $p_j(\cdot \mid \bm{s})$ induced by $P$, the paper uses a theoretically equivalent adaptation of the \emph{streak-distillation} objective defined in Definition \ref{def:sd-objective}. 

Building from this streak-objective, substituting with the equivalency in \Cref{eq:streak-equiv-b} yields,
\begin{align}
    \;\;
    \tfrac{\text{Tokens}}{\text{Draft}}(\bm{x}, \bm{s})
    \;=
    \sum_{m=1}^{\gamma}\;\prod_{j=1}^{m} p_j\;\!\big({x}_j\mid \bm{x}_{t < j}\big),
    \label{eq:tau-rank}
\end{align} 
where $\bm{x} \sim Q^{\text{diff}}(\cdot|\bm{s})$ is now a draft generated by the diffusion model $Q^{\text{diff}}$. This alternate form allows us to determine the expected accepted tokens yielded from a draft $\bm{x}$ using directly the verifier $P$.

Importantly, \Cref{eq:tau-rank} formalizes the degree to which draft $\bm{x}$ is expected to contribute to inference throughput. Thus, for $K$ drafts sampled from $Q^{\text{diff}}(\cdot|\bm{s})$, the throughput-maximizing draft, denoted $\bm{x}^{\max}$, can be selected by,
\begin{equation*}
    \bm{x}^{\max} \triangleq \arg\max_k \left(\tfrac{\text{Tokens}}{\text{Draft}}(\bm{x}^1, \bm{s}\right), \ldots,\tfrac{\text{Tokens}}{\text{Draft}}(\bm{x}^K, \bm{s})),
\end{equation*}
for a shared prefix $\bm{s}$. 
This approach maximizes expected speed-up by construction, selecting based on the expected throughput of each generated draft. 

Operationally, the verifier scores each candidate using token-wise posteriors
$\big\{p_j\big(x^{k}_j \!\mid\! \bm{s} \circ \bm{x}^{k}_{<j}\big)\,\big\}_{j=1}^{\gamma}$,
then selects $\bm{x}^{\max}$ before applying the acceptance rule, yielding the \emph{self-selection} mechanism. As detailed in App.~\ref{A:experimental-details-tt}, these scores across all $K$ drafts can be computed efficiently via tree-style attention~\citep{xiong2024dyspecfasterspeculativedecoding}. A pseudo-code of the \textit{self-selection acceptance} mechanism is provided in Algorithm \ref{alg:self-select}.

\textbf{Lossless verification with greedy-acceptance.} Once a draft $\bm{x}^{\max}$ has been selected by $P$, it then must be verified to determine which tokens to accept and reject. Deploying the greedy-acceptance rule then becomes trivial, given that verifier probabilities over $\bm{x}^{\max}$ are cached form the previous ranking phase. Additionally, note that the drafting probabilities for $\bm{x}^{\max}$ are never considered, enabling cross-tokenizer support and extensions to diffusion drafters that do not output calibrated probability distributions \citep{sahoo2024simple, shi2024simplified}.

\section{Experimental Settings}
\label{sec:experiments}

\begin{figure*}[t]
  \centering
    \includegraphics[width=.99\linewidth]{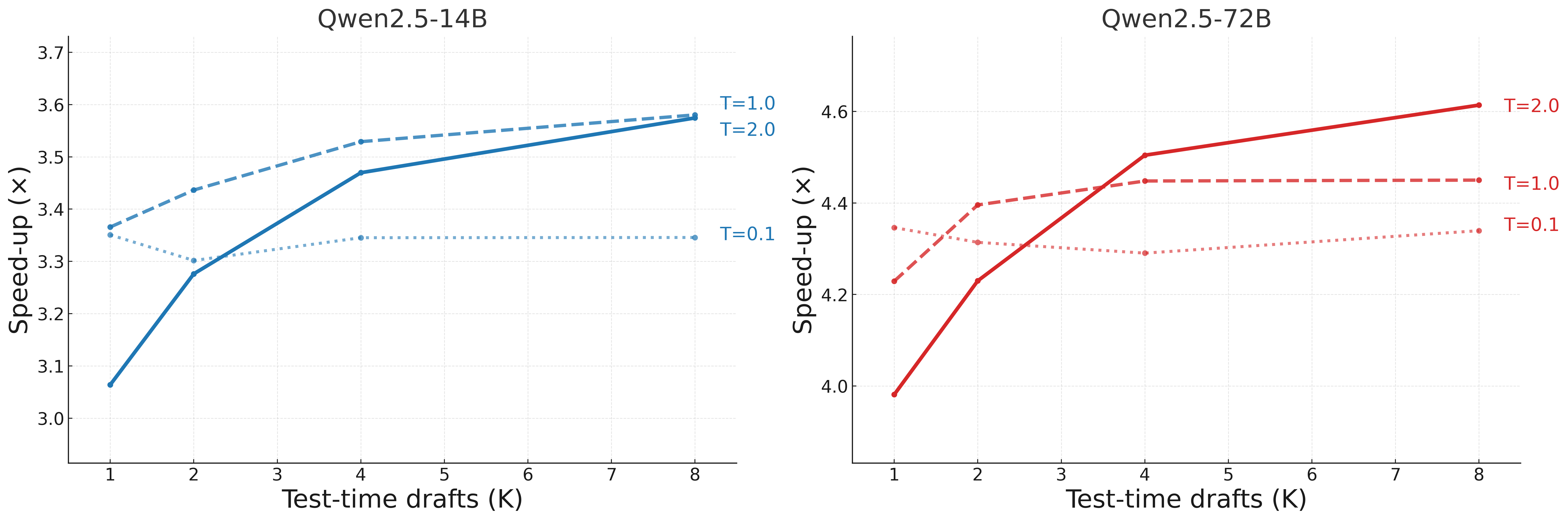}
    \vspace{-5pt}
  \caption{Test-time speed-up scaling w.r.t.~parallel drafts $K$ when deploying self-selection. Tested across Qwen2.5-14B-Instruct, and Qwen2.5-72B-Instruct on Math500 prompts (out of distribution). Tested across different drafter temperatures $T$ above. Showing the positive effect of additional draft variance at large $K$.}
  \label{fig:ssa-scaling}
\end{figure*}

This section presents a comprehensive evaluation of \textsc{SpecDiff-2} against state-of-the-art speculative decoding methods, spanning both autoregressive (AR) and diffusion-based drafters. We study four drafter–verifier pairs across three datasets and report \emph{end-to-end wall-clock speedups}, \emph{token-level acceptance}, and 
\emph{average accepted streak length} under matched decoding budgets.
Our goals are twofold: {\bf (i)} quantify the gains from our test- and train-time alignment in speculative decoding, and {\bf (ii)} characterize when diffusion-based drafting can serve to scale efficient reasoning on low budgets. 
All evaluation keep the output lossless (identical to the verifier model) across all methods. 

\textbf{Models and datasets.}
The evaluation compares \textsc{SpecDiff-2} to strong, publicly released baselines: 
{\bf (i)} \emph{Speculative Sampling (SpS)} \cite{leviathan2023fast} (classical AR draft-then-verify with a smaller AR drafter), 
{\bf (ii)} \emph{EAGLE}, and {\bf (iii)} \emph{EAGLE-2} \cite{li2024eagle,li2024eagle2} (AR drafters with verifier-aligned early-accept mechanisms). These algorithms represent the current state-of-the-art in AR-based speculative decoding. 
Finally, {\bf (iv)} the evaluation includes the original \emph{SpecDiff} \cite{christopher2025speculative} (which uses a unaligned diffusion drafter). SpecDiff is the previous state-of-the-art speculative diffusion method, and it is also used to isolate the impact of the proposed train- and test-time alignment methods.

The evaluation covers three settings that stress different acceptance regimes: 
{\bf (i)} \emph{question answering} (long-form, open-ended reasoning using the \textit{GPQA} dataset \cite{rein2024gpqa}), 
{\bf (ii)} \emph{mathematical reasoning} (which has sparse-support, and requires high-precision acceptance, using dataset \textit{Math-500} \cite{hendrycks2021math}), 
and {\bf (iii)} \emph{code generation} (structured outputs with exactness constraints, using \textit{HumanEval} \cite{chen2021evaluating}). 
These domains jointly probe {\bf (a)} calibration of drafter marginals, {\bf (b)} robustness of acceptance under distribution shift, and {\bf (c)} downstream faithfulness when drafts are partially rejected. All methods use identical prompts, stopping criteria, and verifier decoding parameters per setting. Further, 
wall-clock is measured using A100 80GB GPUs. \emph{Importantly, note that all tested benchmarks lie outside the training/finetuning distribution} for all reported results, thus showcasing strong generalization capabilities of SpecDiff-2.

\begin{table}[t]
\centering
\renewcommand{\arraystretch}{1.15}
\resizebox{\columnwidth}{!}{
\begin{tabular}{l l l}
\toprule
\textbf{Verifier} & \textbf{Drafter} & \textbf{Tokenizer} \\
\midrule
LLaMA-2-70B-chat   & \multirow{2}{*}{DiffuLLaMA-7B} & \multirow{2}{*}{LLaMA-2} \\
LLaMA-2-13B-chat   &                                  &    \\
\hline
Qwen2.5-72B-Instruct & \multirow{2}{*}{DiffuCoder-7B} & \multirow{2}{*}{Qwen2} \\
Qwen2.5-14B-Instruct &                                 &                                   \\[2pt]
\bottomrule
\end{tabular}
}
\caption{Verifier–drafter pairings and shared tokenizers.}
\label{tab:drafters-verifiers}
\vspace{-10pt}
\end{table}

\textbf{Verifiers.} The evaluation selects verifiers to satisfy {\bf (i)} compatibility with high-quality diffusion-LM tokenization schemes and {\bf (ii)} availability of stable, open implementations of state-of-the-art speculative decoders: it uses \textsc{Qwen2.5-72B-Instruct} \cite{qwen2.5} and \textsc{LLaMA-2-70B-chat} \cite{touvron2023llama}, which align with released EAGLE/EAGLE-2 toolchains and tokenizers. Appendix ~\ref{A:extra_experiments} also reports additional results on smaller verifiers (\textsc{Qwen2.5-14B-Instruct} and \textsc{LLaMA-2-13B-chat}).

\textbf{Drafters.} For diffusion drafting, the settings adopts adapted Diffusion Language Models (DLMs) with tokenizer alignment to the chosen verifiers: \textsc{DiffuCoder-7B} (Qwen2.5 tokenizer) \cite{gong2025diffucoder} and \textsc{DiffuLLaMA-7B} (LLaMA-2 tokenizer) \cite{gong2024scaling}. Notably these diffusion drafters are pretrained, thus they enable efficient application of Streak-distillation as \emph{purely a finetuning paradigm.}
For AR drafting baselines, \emph{SpS} uses \textsc{Qwen2.5-7B-Instruct} and \textsc{LLaMA-2-7B-chat} as the small draft models. 
The verifier–drafter pairings and the associated shared tokenizers are summarized in Table~\ref{tab:drafters-verifiers}. 
Note that we rely on the EAGLE and EAGLE-2 implementations for comparability.\footnote{Despite our best attempts, the results from the original EAGLE-3 work were not reproducible in our environment.}


\noindent\textbf{Implementation notes.} 
Details of the drafter hyperparameters and ablations are reported in \Cref{A:experimental-details}, with key details highlighted here to contextualize the evaluation:
\begin{itemize}[leftmargin=*, parsep=0pt, itemsep=0pt, topsep=0pt]
    \item \textbf{Diffusion Steps:} Given the unconventionally large size of our drafter models ($\sim$$7\times$ larger than EAGLE's drafters), it is not neither empirically optimal nor necessary to iterate on drafts for multiple diffusion steps. The diminishing returns of additional diffusion steps is illustrated in Appendix \ref{A:extra_experiments-steps}, which provided only slightly better predictions while scaling draft time linearly.
    \item \textbf{Drafter Temperature:} It is also observed that the drafting temperature can be optimized to balance draft quality (suffering when temperature is high) and sufficient variance among the $K$ drafts to take full advantage of self-selection (suffering when temperature is low); as shown in Figure \ref{fig:ssa-scaling}, $\operatorname{temperature}=1.5$ serves as a consistent middle ground between the two extremes.
\end{itemize}

Beyond these hyperparameters, the study defers over-optimization of the the drafter mechanics, to avoid over-fitting to specific settings, and, consequentially, presenting results that are not robust.

\begin{table*}[ht!]
\centering
\footnotesize
\setlength{\tabcolsep}{4.8pt}
\renewcommand{\arraystretch}{1.15}
\resizebox{\textwidth}{!}{
\begin{tabular}{l l *{3}{cc} cc}
\toprule
\multicolumn{2}{c}{ } &
\multicolumn{2}{c}{\textbf{Math-500}} &
\multicolumn{2}{c}{\textbf{HumanEval}} &
\multicolumn{2}{c}{\textbf{GPQA}} &
\multicolumn{2}{c}{\textbf{Mean}} \\
\cmidrule(lr){3-4}\cmidrule(lr){5-6}\cmidrule(lr){7-8}\cmidrule(lr){9-10}
\textbf{Model} & \textbf{Accelerator} &
\textbf{Speed-up} & $\streak$ &
\textbf{Speed-up} & $\streak$ &
\textbf{Speed-up} & $\streak$ &
\textbf{Speed-up} & $\streak$ \\
\midrule
\multicolumn{10}{c}{\textbf{Temperature = 0}} \\
\midrule
\multirow{4}{*}{Qwen-2.5-72B} 
 & SpS      & 1.87$\times$ & 1.77 toks & 1.75$\times$ & 1.77 toks & 1.47$\times$   & 1.78 toks & 1.70$\times$   & 1.77 toks \\
 & EAGLE    & 2.50$\times$ & 3.78 toks & 2.28$\times$ & 3.34 toks & 1.93\(\times\) & 2.79 toks & 2.24\(\times\) & 3.30 toks \\
 & EAGLE-2  & 3.16$\times$ & 4.69 toks & 3.16$\times$ & 4.87 toks & 2.50\(\times\) & 3.67 toks & 2.94\(\times\) & 4.41 toks \\
& \shad{\textbf{SpecDiff-2}} &\shad{\textbf{ 4.62$\times$}} & \shad{\textbf{6.47 toks}} & \shad{\textbf{4.98$\times$}} & \shad{\textbf{6.98 toks}} & \shad{\textbf{3.28$\times$}} &\shad{\textbf{ 4.59 toks}} & \shad{\textbf{4.29$\times$}} & \shad{\textbf{5.98 toks}} \\
\addlinespace[2pt]
\cline{2-10}
\addlinespace[2pt]
\multirow{4}{*}{LLaMA-2-70B}
 & SpS      & 1.32$\times$ & 1.72 toks & 1.39$\times$ & 1.69 toks & 1.15$\times$   & 1.70 toks & 1.29$\times$   & 1.70 toks \\
 & EAGLE    & 3.00$\times$ & 4.03 toks & 3.11$\times$ & 4.28 toks & 2.47\(\times\) & 3.34 toks & 2.86\(\times\) & 3.88 toks \\
 & EAGLE-2  & 3.48$\times$ & 4.69 toks & 3.87$\times$ & 5.29 toks & 3.08\(\times\) & 4.18 toks & 3.48\(\times\) & 4.72 toks \\
 & \shad{\textbf{SpecDiff-2}} & \shad{\textbf{3.61$\times$}} & \shad{\textbf{5.04 toks}} & \shad{\textbf{4.69$\times$}} & \shad{\textbf{6.57 toks}} & \shad{\textbf{3.48$\times$}} & \shad{\textbf{4.87 toks}} & \shad{\textbf{3.93$\times$}} & \shad{\textbf{5.49 toks}} \\
\midrule
\multicolumn{10}{c}{\textbf{Temperature = 1}} \\
\midrule
\multirow{4}{*}{Qwen-2.5-72B} 
 & SpS      & 1.70$\times$ & 1.87 toks & 1.78$\times$ & 1.96 toks & 1.49$\times$   & 1.98 toks & 1.66$\times$   & 1.94 toks \\
 & EAGLE    & 2.19$\times$ & 3.38 toks & 2.16\(\times\) & 3.20 toks & 1.69\(\times\) & 2.50 toks & 2.01\(\times\) & 3.03 toks \\
 & EAGLE-2  & 3.12$\times$ & 4.63 toks & 3.15\(\times\) & 4.77 toks & 2.52\(\times\) & 3.67 toks & 2.93\(\times\) & 4.36 toks \\
 & \shad{\textbf{SpecDiff-2}} & \shad{\textbf{5.01$\times$}} & \shad{\textbf{7.00 toks}}  & \shad{\textbf{5.51$\times$}} & \shad{\textbf{7.71 toks}} & \shad{\textbf{2.65$\times$}} & \shad{\textbf{3.71 toks}} & \shad{\textbf{4.39$\times$}} & \shad{\textbf{6.14 toks}} \\
\addlinespace[2pt]
\cline{2-10}
\addlinespace[2pt]
\multirow{4}{*}{LLaMA-2-70B} 
 & SpS      & 1.39$\times$ & 1.79 toks & 1.39$\times$ & 1.76 toks & 1.24$\times$   & 1.76 toks & 1.34$\times$   & 1.77 toks \\
 & EAGLE    & 2.81$\times$ & 3.80 toks & 3.22\(\times\) & 4.38 toks & 2.55\(\times\) & 3.38 toks & 2.86\(\times\) & 3.85 toks \\
 & EAGLE-2  & 3.56$\times$& 4.75 toks & 3.91\(\times\) & 5.33 toks & 3.08\(\times\) & 4.20 toks & 3.52\(\times\) & 4.76 toks \\
 & \shad{\textbf{SpecDiff-2}} & \shad{\textbf{3.99$\times$}} & \shad{\textbf{5.58 toks}} & \shad{\textbf{5.28$\times$}} & \shad{\textbf{7.40 toks}} & \shad{\textbf{3.54$\times$}} & \shad{\textbf{4.95 toks}} & \shad{\textbf{4.27$\times$}} & \shad{\textbf{5.98 toks}} \\
\bottomrule
\end{tabular}
}
\caption{Comparison on \textit{Math-500}, \textit{livecodebench}, and \textit{MT-Bench} for four base models at temperatures 0 and 1. Each cell reports relative \textit{Speed-up} and average acceptance length $\frac{\text{Tokens}}{\text{Draft}}$ (where +1 is added to represent token from verifier, consistent with previous literature). Experiments utilize two A100 GPUs (80Gb). } 
\label{tab:EAGLE2-short-style-qwen-llama}
\vspace{-13pt}
\end{table*}

\section{Experimental Evaluation}
\label{sec:results}

\subsection{Wall-Clock Speedups and Accepted Streaks}

To assess the performance of different speculative decoders, the speed-up is reported as compared to vanilla generation with the verifier, alongside the average accepted streak $\tfrac{\text{Tokens}}{\text{Draft}}(\bm x, \bm s)$. 
Other text quality metrics (e.g., PPL, ROUGE, BLEU) are omitted, as the lossless decoding scheme results in outputs that match the performance of the verifier model exactly. The results are reported in \Cref{tab:EAGLE2-short-style-qwen-llama}. 

Note that \textit{SpecDiff-2} consistently achieves the highest speed-ups and longest accept streaks, outperforming state-of-the-art models, and reporting an average \textbf{4.22\(\times\) speed-up} across all settings, increasing from EAGLE-2 by over 30\%. 
Furthermore, when applied to settings where the drafter specialize, such as when \textsc{DiffuCoder} is applied for drafting on coding questions, speed-ups soar to over \textbf{5\(\times\) faster} than autoregressive generation.

The table highlights two key trends. First, larger-capacity drafters (7B vs.~$\sim$1B for EAGLE-2) yield substantially higher $\tfrac{\text{Tokens}}{\text{Draft}}$, and, as expected, these longer accepted streaks translate directly into larger speed-ups. 
Second, because diffusion drafting is natively parallel, the draft latency remains comparable to EAGLE/EAGLE-2 despite their smaller drafters, which require multiple forward passes to produce the same sequence.

\textit{SpecDiff-2} realizes the highest speed-ups when operating on math and coding questions, reporting an average \textbf{4.71\(\times\) speed-up} across verifiers and temperatures compared to EAGLE-2's 3.43\(\times\) speed-up.
In more open-ended QA, the margin tightens to 3.24\(\times\) for \textit{SpecDiff-2} and 2.80\(\times\) for EAGLE-2. 
While both methods slow on open-ended generation where semantic diversity is higher, \textit{SpecDiff-2} still improves over the state of the art, with relative gains tapering from (37\%) (reasoning) to (16\%) (Q\&A).
These results suggest that diffusion drafters excel when the target distribution is more structurally constrained (e.g., code, stepwise reasoning), \emph{motivating \textit{SpecDiff-2} specifically for accelerating structured LLM reasoning}.

\begin{figure*}[t]
  \centering
    \includegraphics[width=.99\textwidth]{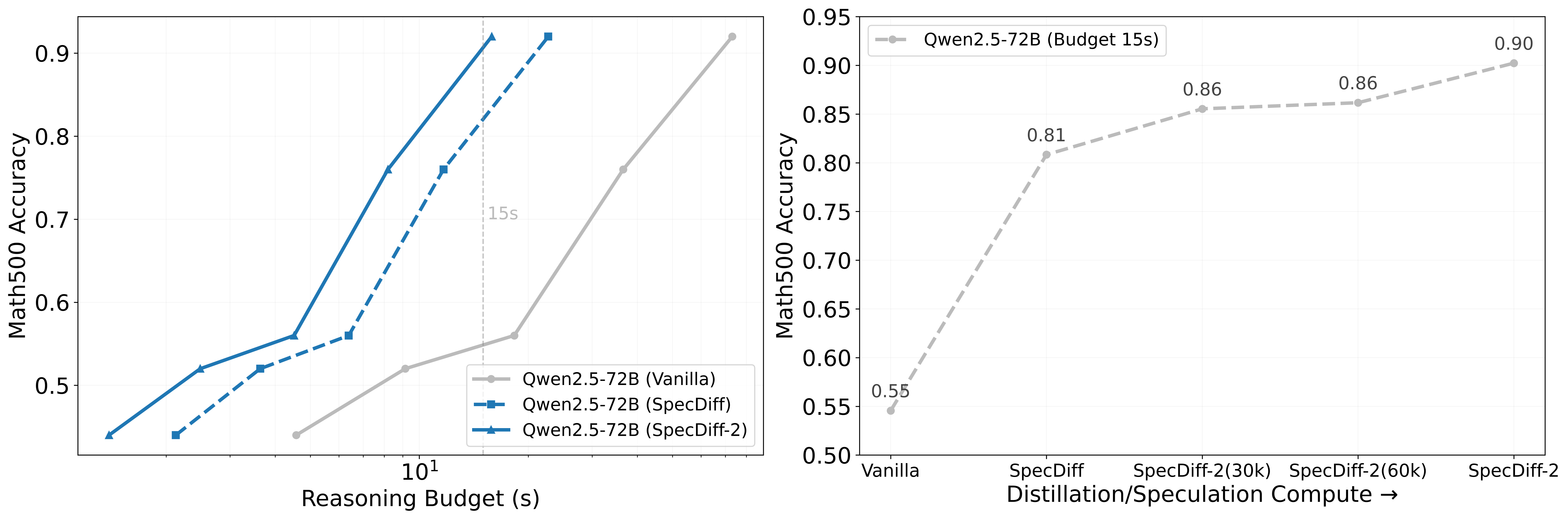}
    \vspace{-10pt}
  \caption{Test-time scaling with respect to wall-time reasoning budget with CoT style prompting. Additional budget results in greater accuracy, accelerated model [Qwen2.5-72B-(SpecDiff-2) above] exhibits $+63\%$ accuracy increase over vanilla model [Qwen2.5-72B(Vanilla) above], and $+11\%$ accuracy over Qwen2.5-72B-(SpecDiff) under identical wall-time constraints (15s). Checkpoints at (30k) and (60k) distillation steps are also shown.}
  \label{fig:tt-scaling}
\end{figure*}

\subsection{Scaling Efficient Reasoning With SpecDiff-2} 
This paper argues that a significant motivation for the development of inference acceleration algorithms is addressing the run-time latency introduced with the emergence of test-time compute scaling.

As shown in several recent works \cite{tian2025think, eisenstadt2025overclocking, zhang2025survey} additional wall-time spent while ``thinking'' at test-time produces increasing accuracy on downstream tasks. In time-constrained settings, however, accuracy is bounded by throughput: faster decoding converts the same wall-time into more usable reasoning tokens. Here, we quantify that link and shows how acceleration from \textit{SpecDiff-2} compounds with test-time compute to raise task accuracy under fixed budgets. 

To show this we use Qwen2.5-72B-Instruct \citep{qwen2.5} (a post-trained model with strong instruction following capabilities). Provided a prompt from the Math500 benchmark, the model is instructed to think through its answers before responding in a chain of thought (CoT) style. After a budget of $b$ seconds has been reached, the prompt:
\begin{quote}
\vspace{-5pt}
\textit{``thinking time is up, wrap up your answer'' }
\vspace{-5pt}
\end{quote}
is appended, and the correctness of the final answer is evaluated (see Appendix \ref{A:example_completions} for example completions). 

\Cref{fig:tt-scaling} (left) shows that increasing the reasoning budget (x-axis) directly influences the accuracy on Math500 prompts, where as expected, more reasoning time allows for greater accuracy. 
Consequently, because \textit{SpecDiff-2} increases throughput, the accelerated system attains higher accuracy at the same wall-time. 
In \Cref{fig:tt-scaling} (right), the accelerated version of Qwen2.5-72B experiences a $\mathbf{+63\%}$ boost in accuracy over the base model, when restricted to think for $b=15$ seconds, and a further ($\mathbf{+11\%}$) over unaligned SpecDiff at the same budget.

A clear relationship emerges between the additional compute spent on acceleration as well as alignment, and the accuracy of the system under fixed wall-time constraints. As shown in \Cref{fig:tt-scaling} (right), increasing alignment/training compute (e.g., streak-distillation steps from \texttt{30k} to \texttt{60k}), adding test-time self-selection, and using speculative diffusion each raise acceptance and effective parallelism, which translates into higher accuracy within the same budget. 
This perspective positions \emph{acceleration compute} as a practical scaling knob: investing more in alignment and fast drafting yields monotonic improvements in time-limited reasoning, making \textit{SpecDiff-2} particularly attractive for structured, CoT-heavy workloads. The paper therefore presents `acceleration-compute' scaling as a \emph{unique and novel scaling paradigm}, enabling a new axis along which model performance can be scaled.

\subsection{Ablations: Train-time and Test-time Scaling}
\label{subsec:ablations}

In \Cref{tab:math500-speed-up}, the baseline SpecDiff provides a qualitative comparison of diffusion drafters in the absence of the train-time and test-time techniques introduced by this paper. Next, these contributions are isolated to assess their individual attribution to the reported speed-ups.

\begin{table}[ht!]
\centering
\begin{tabular}{l l c}
\toprule
\textbf{Model} & \textbf{Accelerator} & \textbf{Speed-Up} \\
\midrule
\multirow{2}{*}{Qwen2.5-72B} & SpecDiff   & 3.19$\times\;\;\;\;\;\;\;\;\;\;\;\;\;\;\;\;$ \\
                             & \shad{\textbf{SpecDiff-2}} &  \shad{\textbf{4.62}$\times$ ($+44.8\%$)}\\
\multirow{2}{*}{Qwen2.5-14B} & SpecDiff   & 2.51$\times\;\;\;\;\;\;\;\;\;\;\;\;\;\;\;\;$ \\
                             & \shad{\textbf{SpecDiff-2}} & \shad{\textbf{3.65}$\times$ ($+45.5\%$)} \\
\bottomrule
\end{tabular}
\caption{Math500 comparison for SpecDiff (unaligned drafter) and SpecDiff-2 (this work), with greedy decoding.}
\label{tab:math500-speed-up}
\end{table}

\begin{figure}[t]
  \begin{subfigure}[t]{1.0\linewidth}
    \includegraphics[height=2.2in]{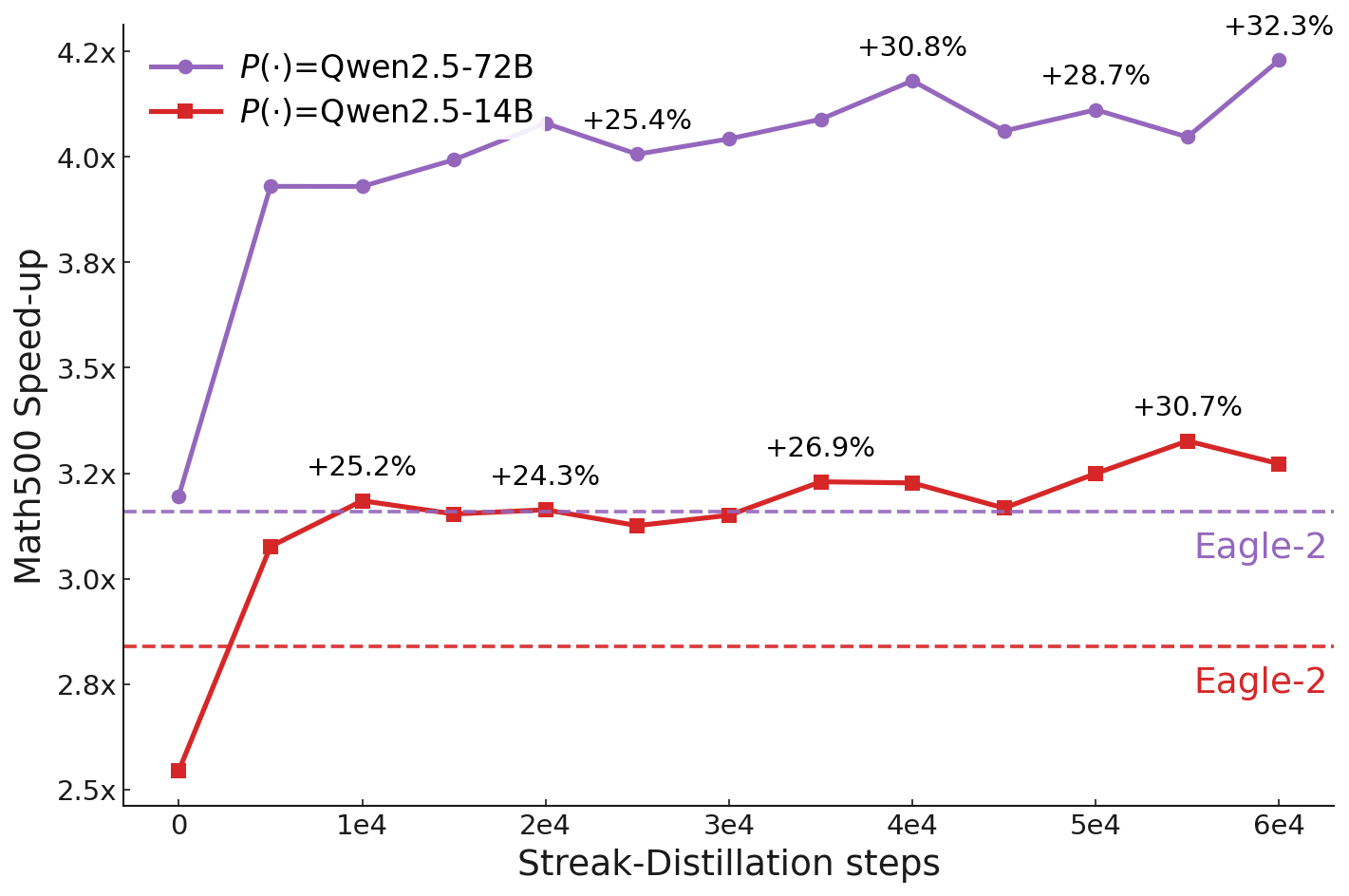}
  \end{subfigure}
  \vspace{-5pt}
  \caption{Scaling Streak-distillation for larger scale drafter alignment on reasoning corpus (see Appendix \ref{A:experimental-details} for details). We see $\sim 30\%$ increases in acceleration across out-of-distribution Math500 prompts, surpassing SOTA baselines (Eagle-2) in the process. $P(\cdot)$ shows verifier model, \textcolor{purple}{red} is Qwen2.5-14B-Instruct (Qwen2.5-14B above), \textcolor{violet}{purple} is Qwen2.5-72B-Instruct (Qwen2.5-72B above), with corresponding Eagle-2 baselines.} 
  \label{fig:sd-full}
\end{figure}

The \textbf{train-time scaling} curve in \Cref{fig:sd-full}, illustrates the increased throughput scaling with respect to \textit{streak-distillation} steps. Validating over the first 60,000 train-steps, the figure showcases the efficacy of streak-distillation in improving throughput with minimal train-time compute, $(\leq 75)$ GPU-hours (see Appendix \ref{A:experimental-details} for hardware details). Both Qwen models see a $\sim30\%$ increase in speed-up, comparable to improvements seen in autoregressive drafters from DistillSpec \cite{zhou2023distillspec}. Ultimately, the distilled drafters exceed both EAGLE-2 baseline throughput's, by up-to $+32\%$.

Next, the ablation investigates empirical \textbf{test-time scaling} trends associated with the novel acceptance scheme \emph{self-selection}. \emph{Self-selection} is tested on Math500 \citep{hendrycks2021math} prompts across both Qwen2.5 models (see Appendix \ref{A:extra_experiments-llama} for LLaMA scaling). The drafting temperature is varied, (`$\text{T}$' in Figure \ref{fig:ssa-scaling}), and the number of drafts is scaled $K = 1, \ldots, 8$, measuring the speed-up at each stage. The results illustrate smooth test-time scaling in speed-up for all model pairs, where a larger number of drafts $K$ increases the probability that the verifier can select an aligned sample, as intended. The largest gains appear at $T=2.0$, reaching up to $+20\%$ additional throughput at $K=8$. By contrast, low temperatures $T=0.1$ show little scaling, reflecting low variance across drafts.

These results show that the proposed train-time distillation provides on average a +30\% increase in the overall speed-up, while the test-time algorithm pushes the speed-up by an additional +15\%. This effectively results in 40-50\% performance improvement over the unaligned SpecDiff. 

\section{Related Work}

Early work in LLM inference acceleration focused primarily on architectural changes, such as model pruning \citep{sun2024simpleeffectivepruningapproach} and quantization \cite{frantar2023gptqaccurateposttrainingquantization}, enabling more efficient generation at the cost of overall generation quality. Addressing this, speculative decoding emerged as a promising alternative, accelerating generation without degrading the generations \cite{leviathan2023fast, chen2023accelerating}. While broadly regarded as a trans formative contribution, however, early works presented moderate speed-ups, motivating further exploration of speculative drafting techniques.
Existing literature on the topic can be partitioned into two categories, with prior works addressing either the autoregressive dependency during drafting or the misalignment between the draft and verifier models (e.g., one of the two \emph{speculative-bottlenecks} reviewed in the Introduction section.

\textbf{Draft model alignment} has been improved through both specialized training procedures and test-time drafting enhancements.
In particular, several approaches build on principles from knowledge distillation \cite{zhou2023distillspec} and optimal transport \cite{sun2023spectr} to design training-time objectives that align the draft model distributions more directly with the verifier posterior. These approaches, however, are specifically catered to AR drafting architectures. Another line of work has realized significant success by implementing clever drafting algorithms, such as employing draft-trees, which generally result in much higher throughput per draft-then-verify step \citep{li2024eagle, li2024eagle2, li2025eagle3}. This work has compared against such methods. 

\textbf{Draft model latency} has been studied separately, leveraging draft models with parallel draft heads \citep{ankner2024hydra}. 
More recent work demonstrated that non-autoregressive diffusion language models can operate as highly efficient drafters, similarly removing sequential dependencies from the drafting step \cite{christopher2025speculative}. 

Our analysis reports substantial improvements against these proposals.

\section{Future Work and Limitations}

We conclude by outlining the main limitations of \emph{SpecDiff-2} and the most promising directions they expose. At a high level, the presented results broaden the design space of acceptance rules and drafting strategies for speculative decoding, but several systems and theory questions araise.

First, note that acceptance in speculative decoding is typically fixed. The \emph{greedy-acceptance} rule proposed in this paper shows that alternative criteria can be both practical and advantageous. Formal analysis of these rules (e.g., their bias/variance trade-offs, worst-case acceptance under miscalibration, and interactions with verifier temperature) remain an open direction. Because greedy-acceptance naturally supports heterogeneous drafters and mismatched tokenizers, a systematic study of \emph{cross-family} and \emph{cross-tokenizer} drafting (rather than same-family pairs) is a key next step.

Beyond tokenization schemes, the proposed self-selection drafting mechanism sets the stage for different drafting procedure producing multiple drafts in parallel. A natural extension is to deploy $K$ \emph{distinct} drafters specialized to sub-distributions (e.g., algebraic manipulation vs.\ numeric computation vs.\ code) and to select among them at test time. Doing so raises open questions about diversity–alignment trade-offs, cost-aware selection policies, and how best to allocate a fixed compute budget across $K$, temperature, and proposal depth.

This work also introduces an important question of the optimal size for diffusion draft models. While this has been well established for autoregressive architectures, the empirical results show that the parallel drafting properties of diffusion enable scaling to much larger drafter architectures, while maintaining similar drafter latency.
In this work, drafter selection was largely determined by diffusion models quality, with \textsc{DiffuLLaMA} and \textsc{DiffuCoder} being some of the strongest open-source DLMs, but the lower speed-ups for smaller verifiers suggest that greater speed-ups could be realized by scaling the DLM drafters proportionally to the verifiers. Deriving scaling laws and compute–latency frontiers for diffusion drafters, ideally as functions of $\gamma$, acceptance, and verifier size, remains an open direction.

Finally, there is significant room for hardware based optimization. Efficient kernels specialized for semi-autoregressive inference are lacking, and support for KV caching in the context of $\gamma$ length diffusion drafts is important for efficient inference.

\section{Conclusion}

This study has presented \emph{SpecDiff-2}, providing the first work on model alignment for speculative decoding with diffusion draft models. Motivated by the absence of approaches which simultaneously address drafter latency and drafter-verifier alignment, this paper introduces train-time distillation techniques and test-time parallel draft generation catered for diffusion drafters.
Leveraging these novel diffusion alignment methods, the proposed framework observes significant increase in token throughput and achieves up to an average of $+55\%$ faster generation over previous baselines and up to $5.5\times$ average speed-up over standard decoding, without any loss of accuracy.
\textit{SpecDiff-2} achieves state-of-the-art performance across a breadth of reasoning, coding, and mathematical benchmarks, establishing a new standard for lossless acceleration. 

We believe that the use of diffusion drafter represents a significant step towards inference-time acceleration, opening several promising directions for accelerating LLMs.

\section*{Acknowledgments}
This work was partially supported by an NVIDIA Academic Grant Program Award. 
The work was also supported by NSF grants 2334448, 2401285, and 2533631, DARPA Contracting activity
 \#HR0011252E005, and UVA’s National Security Data \& Policy Institute, ODNI Contracting Activity \#2024-24070100001. 
Its view and conclusions reflect those of
the authors only.

\bibliography{iclr2026_conference}

\begin{thebibliography}{29}
\providecommand{\natexlab}[1]{#1}
\providecommand{\url}[1]{\texttt{#1}}
\expandafter\ifx\csname urlstyle\endcsname\relax
  \providecommand{\doi}[1]{doi: #1}\else
  \providecommand{\doi}{doi: \begingroup \urlstyle{rm}\Url}\fi

\bibitem[Ankner et~al.(2024)Ankner, Parthasarathy, Nrusimha, Rinard, Ragan-Kelley, and Brandon]{ankner2024hydra}
Ankner, Z., Parthasarathy, R., Nrusimha, A., Rinard, C., Ragan-Kelley, J., and Brandon, W.
\newblock Hydra: Sequentially-dependent draft heads for medusa decoding.
\newblock \emph{arXiv preprint arXiv:2402.05109}, 2024.

\bibitem[Chen et~al.(2023)Chen, Borgeaud, Irving, Lespiau, Sifre, and Jumper]{chen2023accelerating}
Chen, C., Borgeaud, S., Irving, G., Lespiau, J.-B., Sifre, L., and Jumper, J.
\newblock Accelerating large language model decoding with speculative sampling.
\newblock \emph{arXiv preprint arXiv:2302.01318}, 2023.

\bibitem[Chen et~al.(2021)Chen, Tworek, Jun, Yuan, Pinto, Kaplan, Edwards, Burda, Joseph, Brockman, et~al.]{chen2021evaluating}
Chen, M., Tworek, J., Jun, H., Yuan, Q., Pinto, H. P. D.~O., Kaplan, J., Edwards, H., Burda, Y., Joseph, N., Brockman, G., et~al.
\newblock Evaluating large language models trained on code.
\newblock \emph{arXiv preprint arXiv:2107.03374}, 2021.

\bibitem[Christopher et~al.(2025)Christopher, Bartoldson, Ben-Nun, Cardei, Kailkhura, and Fioretto]{christopher2025speculative}
Christopher, J.~K., Bartoldson, B.~R., Ben-Nun, T., Cardei, M., Kailkhura, B., and Fioretto, F.
\newblock Speculative diffusion decoding: Accelerating language generation through diffusion.
\newblock In Chiruzzo, L., Ritter, A., and Wang, L. (eds.), \emph{Proceedings of the 2025 Conference of the Nations of the Americas Chapter of the Association for Computational Linguistics: Human Language Technologies (Volume 1: Long Papers)}, pp.\  12042--12059, Albuquerque, New Mexico, April 2025. Association for Computational Linguistics.
\newblock ISBN 979-8-89176-189-6.
\newblock \doi{10.18653/v1/2025.naacl-long.601}.
\newblock URL \url{https://aclanthology.org/2025.naacl-long.601/}.

\bibitem[Eisenstadt et~al.(2025)Eisenstadt, Zimerman, and Wolf]{eisenstadt2025overclocking}
Eisenstadt, R., Zimerman, I., and Wolf, L.
\newblock Overclocking llm reasoning: Monitoring and controlling thinking path lengths in llms.
\newblock \emph{arXiv preprint arXiv:2506.07240}, 2025.

\bibitem[Frantar et~al.(2023)Frantar, Ashkboos, Hoefler, and Alistarh]{frantar2023gptqaccurateposttrainingquantization}
Frantar, E., Ashkboos, S., Hoefler, T., and Alistarh, D.
\newblock Gptq: Accurate post-training quantization for generative pre-trained transformers, 2023.
\newblock URL \url{https://arxiv.org/abs/2210.17323}.

\bibitem[Gong et~al.(2024)Gong, Agarwal, Zhang, Ye, Zheng, Li, An, Zhao, Bi, Han, et~al.]{gong2024scaling}
Gong, S., Agarwal, S., Zhang, Y., Ye, J., Zheng, L., Li, M., An, C., Zhao, P., Bi, W., Han, J., et~al.
\newblock Scaling diffusion language models via adaptation from autoregressive models.
\newblock \emph{arXiv preprint arXiv:2410.17891}, 2024.

\bibitem[Gong et~al.(2025)Gong, Zhang, Zheng, Gu, Jaitly, Kong, and Zhang]{gong2025diffucoder}
Gong, S., Zhang, R., Zheng, H., Gu, J., Jaitly, N., Kong, L., and Zhang, Y.
\newblock Diffucoder: Understanding and improving masked diffusion models for code generation.
\newblock \emph{arXiv preprint arXiv:2506.20639}, 2025.

\bibitem[Hendrycks et~al.(2021)Hendrycks, Burns, Basart, Critch, Li, Song, and Steinhardt]{hendrycks2021math}
Hendrycks, D., Burns, C., Basart, S., Critch, A., Li, J., Song, D., and Steinhardt, J.
\newblock Measuring mathematical problem solving with the {MATH} dataset.
\newblock \emph{arXiv preprint arXiv:2103.03874}, 2021.

\bibitem[Leviathan et~al.(2023)Leviathan, Kalman, and Matias]{leviathan2023fast}
Leviathan, Y., Kalman, M., and Matias, Y.
\newblock Fast inference from transformers via speculative decoding.
\newblock In \emph{International Conference on Machine Learning}, pp.\  19274--19286. PMLR, 2023.

\bibitem[Li et~al.(2024{\natexlab{a}})Li, Wei, Zhang, and Zhang]{li2024eagle}
Li, Y., Wei, F., Zhang, C., and Zhang, H.
\newblock Eagle: Speculative sampling requires rethinking feature uncertainty.
\newblock \emph{arXiv preprint arXiv:2401.15077}, 2024{\natexlab{a}}.

\bibitem[Li et~al.(2024{\natexlab{b}})Li, Wei, Zhang, and Zhang]{li2024eagle2}
Li, Y., Wei, F., Zhang, C., and Zhang, H.
\newblock Eagle-2: Faster inference of language models with dynamic draft trees.
\newblock \emph{arXiv preprint arXiv:2406.16858}, 2024{\natexlab{b}}.

\bibitem[Li et~al.(2025)Li, Wei, Zhang, and Zhang]{li2025eagle3}
Li, Y., Wei, F., Zhang, C., and Zhang, H.
\newblock Eagle-3: Scaling up inference acceleration of large language models via training-time test.
\newblock \emph{arXiv preprint arXiv:2503.01840}, 2025.

\bibitem[Rein et~al.(2024)Rein, Hou, Stickland, Petty, Pang, Dirani, Michael, and Bowman]{rein2024gpqa}
Rein, D., Hou, B.~L., Stickland, A.~C., Petty, J., Pang, R.~Y., Dirani, J., Michael, J., and Bowman, S.~R.
\newblock Gpqa: A graduate-level google-proof q\&a benchmark.
\newblock In \emph{First Conference on Language Modeling}, 2024.

\bibitem[Sahoo et~al.(2024)Sahoo, Arriola, Schiff, Gokaslan, Marroquin, Chiu, Rush, and Kuleshov]{sahoo2024simple}
Sahoo, S., Arriola, M., Schiff, Y., Gokaslan, A., Marroquin, E., Chiu, J., Rush, A., and Kuleshov, V.
\newblock Simple and effective masked diffusion language models.
\newblock \emph{Advances in Neural Information Processing Systems}, 37:\penalty0 130136--130184, 2024.

\bibitem[Schick et~al.(2023)Schick, Dwivedi-Yu, Dessi, Raileanu, Lomeli, and Severyn]{schick2023toolformer}
Schick, T., Dwivedi-Yu, K., Dessi, R., Raileanu, R., Lomeli, M., and Severyn, A.
\newblock Toolformer: Language models can teach themselves to use tools.
\newblock \emph{arXiv preprint arXiv:2302.04761}, 2023.

\bibitem[Shi et~al.(2024)Shi, Han, Wang, Doucet, and Titsias]{shi2024simplified}
Shi, J., Han, K., Wang, Z., Doucet, A., and Titsias, M.
\newblock Simplified and generalized masked diffusion for discrete data.
\newblock \emph{Advances in neural information processing systems}, 37:\penalty0 103131--103167, 2024.

\bibitem[Sun et~al.(2024)Sun, Liu, Bair, and Kolter]{sun2024simpleeffectivepruningapproach}
Sun, M., Liu, Z., Bair, A., and Kolter, J.~Z.
\newblock A simple and effective pruning approach for large language models, 2024.
\newblock URL \url{https://arxiv.org/abs/2306.11695}.

\bibitem[Sun et~al.(2023)Sun, Suresh, Ro, Beirami, Jain, and Yu]{sun2023spectr}
Sun, Z., Suresh, A.~T., Ro, J.~H., Beirami, A., Jain, H., and Yu, F.
\newblock Spectr: Fast speculative decoding via optimal transport.
\newblock \emph{Advances in Neural Information Processing Systems}, 36:\penalty0 30222--30242, 2023.

\bibitem[Tian et~al.(2025)Tian, Zhao, Wang, Chen, Ji, Peng, Zhao, and Li]{tian2025think}
Tian, X., Zhao, S., Wang, H., Chen, S., Ji, Y., Peng, Y., Zhao, H., and Li, X.
\newblock Think twice: Enhancing llm reasoning by scaling multi-round test-time thinking.
\newblock \emph{arXiv preprint arXiv:2503.19855}, 2025.

\bibitem[Touvron et~al.(2023)Touvron, Martin, Stone, Albert, Almahairi, Babaei, Bashlykov, Batra, Bhargava, Bhosale, et~al.]{touvron2023llama}
Touvron, H., Martin, L., Stone, K., Albert, P., Almahairi, A., Babaei, Y., Bashlykov, N., Batra, S., Bhargava, P., Bhosale, S., et~al.
\newblock Llama 2: Open foundation and fine-tuned chat models.
\newblock \emph{arXiv preprint arXiv:2307.09288}, 2023.

\bibitem[Wang et~al.(2022)Wang, Wei, Schuurmans, Le, Chi, Narang, Chowdhery, and Zhou]{wang2022self}
Wang, X., Wei, J., Schuurmans, D., Le, Q., Chi, E., Narang, S., Chowdhery, A., and Zhou, D.
\newblock Self-consistency improves chain of thought reasoning in language models.
\newblock \emph{arXiv preprint arXiv:2203.11171}, 2022.

\bibitem[Wei et~al.(2022)Wei, Wang, Schuurmans, Bosma, Xia, Chi, Le, Zhou, et~al.]{wei2022chain}
Wei, J., Wang, X., Schuurmans, D., Bosma, M., Xia, F., Chi, E., Le, Q.~V., Zhou, D., et~al.
\newblock Chain-of-thought prompting elicits reasoning in large language models.
\newblock \emph{Advances in neural information processing systems}, 35:\penalty0 24824--24837, 2022.

\bibitem[Xiong et~al.(2024)Xiong, Zhang, Li, Wu, and Zou]{xiong2024dyspecfasterspeculativedecoding}
Xiong, Y., Zhang, R., Li, Y., Wu, T., and Zou, L.
\newblock Dyspec: Faster speculative decoding with dynamic token tree structure, 2024.
\newblock URL \url{https://arxiv.org/abs/2410.11744}.

\bibitem[Yan et~al.(2024)Yan, Agarwal, and Venkataraman]{yan2024decoding}
Yan, M., Agarwal, S., and Venkataraman, S.
\newblock Decoding speculative decoding.
\newblock \emph{arXiv preprint arXiv:2402.01528}, 2024.

\bibitem[Yang et~al.(2024)Yang, Yang, Zhang, Hui, Zheng, Yu, Li, Liu, Huang, Wei, Lin, Yang, Tu, Zhang, Yang, Yang, Zhou, Lin, Dang, Lu, Bao, Yang, Yu, Li, Xue, Zhang, Zhu, Men, Lin, Li, Tang, Xia, Ren, Ren, Fan, Su, Zhang, Wan, Liu, Cui, Zhang, and Qiu]{qwen2.5}
Yang, A., Yang, B., Zhang, B., Hui, B., Zheng, B., Yu, B., Li, C., Liu, D., Huang, F., Wei, H., Lin, H., Yang, J., Tu, J., Zhang, J., Yang, J., Yang, J., Zhou, J., Lin, J., Dang, K., Lu, K., Bao, K., Yang, K., Yu, L., Li, M., Xue, M., Zhang, P., Zhu, Q., Men, R., Lin, R., Li, T., Tang, T., Xia, T., Ren, X., Ren, X., Fan, Y., Su, Y., Zhang, Y., Wan, Y., Liu, Y., Cui, Z., Zhang, Z., and Qiu, Z.
\newblock Qwen2.5 technical report.
\newblock \emph{arXiv preprint arXiv:2412.15115}, 2024.

\bibitem[Zelikman et~al.(2022)Zelikman, Wu, Mu, and Goodman]{zelikman2022star}
Zelikman, M., Wu, Y., Mu, J., and Goodman, N.~D.
\newblock {STaR}: Bootstrapping reasoning with reasoning.
\newblock \emph{arXiv preprint arXiv:2203.14465}, 2022.

\bibitem[Zhang et~al.(2025)Zhang, Lyu, Sun, Wang, Zhang, Hua, Wu, Guo, Wang, Muennighoff, et~al.]{zhang2025survey}
Zhang, Q., Lyu, F., Sun, Z., Wang, L., Zhang, W., Hua, W., Wu, H., Guo, Z., Wang, Y., Muennighoff, N., et~al.
\newblock A survey on test-time scaling in large language models: What, how, where, and how well?
\newblock \emph{arXiv preprint arXiv:2503.24235}, 2025.

\bibitem[Zhou et~al.(2023)Zhou, Lyu, Rawat, Menon, Rostamizadeh, Kumar, Kagy, and Agarwal]{zhou2023distillspec}
Zhou, Y., Lyu, K., Rawat, A.~S., Menon, A.~K., Rostamizadeh, A., Kumar, S., Kagy, J.-F., and Agarwal, R.
\newblock Distillspec: Improving speculative decoding via knowledge distillation.
\newblock \emph{arXiv preprint arXiv:2310.08461}, 2023.

\end{thebibliography}
\bibliographystyle{mlsys2025}

\newpage
\appendix

\begin{table*}[t!]
\centering
\footnotesize
\setlength{\tabcolsep}{4.8pt}
\renewcommand{\arraystretch}{1.15}
\resizebox{\textwidth}{!}{
\begin{tabular}{l l *{3}{cc} cc}
\toprule
\multicolumn{2}{c}{ } &
\multicolumn{2}{c}{\textbf{Math-500}} &
\multicolumn{2}{c}{\textbf{HumanEval}} &
\multicolumn{2}{c}{\textbf{GPQA}} &
\multicolumn{2}{c}{\textbf{Mean}} \\
\cmidrule(lr){3-4}\cmidrule(lr){5-6}\cmidrule(lr){7-8}\cmidrule(lr){9-10}
\textbf{Model} & \textbf{Accelerator} &
\textbf{Speed-up} & $\streak$ &
\textbf{Speed-up} & $\streak$ &
\textbf{Speed-up} & $\streak$ &
\textbf{Speed-up} & $\streak$ \\
\midrule
\multicolumn{10}{c}{\textbf{Temperature = 0}} \\
\midrule
\multirow{3}{*}{Qwen-2.5-14B} 
 & EAGLE    & 2.59$\times$ & 3.83 toks & 2.57$\times$ & 4.02 toks & 1.94\(\times\) & 2.79 toks & 2.37\(\times\) & 3.55 toks \\
 & EAGLE-2  & 2.84$\times$ & 4.25 toks & 3.05$\times$ & 4.89 toks & \textbf{2.29\(\times\)} & 3.46 toks & 2.73\(\times\) & 4.20 toks \\
 & \shad{\textbf{SpecDiff-2.0}} & \shad{\textbf{3.65$\times$}} & \shad{\textbf{6.20 toks}} & \shad{\textbf{6.17$\times$}} & \shad{\textbf{10.8 toks}} & \shad{2.14$\times$} & \shad{\textbf{3.76 toks}} & \shad{\textbf{3.98$\times$}} & \shad{\textbf{6.92 toks}} \\
\addlinespace[2pt]
\cline{2-10}
\addlinespace[2pt]
\multirow{3}{*}{LLaMA-2-13B} 
 & EAGLE    & 3.43$\times$ & 4.65 toks & 3.47\(\times\) & 4.82 toks & 2.83$\times$ & 3.92 toks & 3.24$\times$ & 4.46 toks \\
 & EAGLE-2  & 3.48$\times$ & 4.83 toks & \textbf{3.92\(\times\)} & 5.63 toks & \textbf{3.04$\times$} & \textbf{4.46 toks} & \textbf{3.48\(\times\)} & 4.97 toks \\
 & \shad{\textbf{SpecDiff-2.0}} & \shad{\textbf{3.61$\times$}} & \shad{\textbf{6.31 toks}} & \shad{3.55$\times$} & \shad{\textbf{6.21 toks}}  & \shad{2.49$\times$} & \shad{4.37 toks} & \shad{3.22$\times$} & \shad{\textbf{5.63 toks}} \\
\midrule
\multicolumn{10}{c}{\textbf{Temperature = 1}} \\
\midrule
\multirow{3}{*}{Qwen-2.5-14B} 
 & EAGLE    & 2.43$\times$ & 3.52 toks & 2.72$\times$ & 4.14 toks & 1.94\(\times\) & 2.81 toks & 2.36\(\times\)  & 3.49 toks \\
 & EAGLE-2  & 3.00$\times$ & 4.32 toks & 3.21$\times$ & 5.03 toks & 2.38\(\times\) & {3.53 toks} & 2.86\(\times\) & 4.29 toks  \\
 & \shad{\textbf{SpecDiff-2.0}} & \shad{\textbf{3.88$\times$}} & \shad{\textbf{5.79 toks}} & \shad{\textbf{3.83$\times$}} & \shad{\textbf{5.69 toks}} & \shad{\textbf{2.53$\times$}} & \shad{\textbf{4.43 toks}} & \shad{\textbf{3.41$\times$}} & \shad{\textbf{5.30 toks}} \\
\addlinespace[2pt]
\cline{2-10}
\addlinespace[2pt]
\multirow{3}{*}{LLaMA-2-13B} 
 & EAGLE    & 3.39$\times$ & 4.63 toks & 3.30\(\times\) & 4.73 toks & 2.77$\times$ & 3.90 toks & 3.15\(\times\) & 4.42 toks \\
 & EAGLE-2  & \textbf{3.43$\times$} & 4.77 toks & \textbf{3.85\(\times\)} & \textbf{5.60 toks} & \textbf{3.14$\times$} & \textbf{4.49 toks} & \textbf{3.47\(\times\)} & 4.95 toks \\
 & \shad{\textbf{SpecDiff-2.0}} & \shad{3.36$\times$} & \shad{\textbf{5.88 toks}}  & \shad{3.15$\times$} & \shad{5.52 toks} & \shad{2.27$\times$} & \shad{3.98 toks} & \shad{2.92$\times$} & \shad{\textbf{5.12 toks}} \\
\bottomrule
\end{tabular}
}
\caption{Comparison on \textit{Math-500}, \textit{livecodebench}, and \textit{MT-Bench} for four base models at temperatures 0 and 1. Each cell reports relative \textit{Speed-up} and average acceptance length $\tau$. Experiments utilize a single Ax100 GPU (80Gb).}
\label{tab:EAGLE2-style-qwen-llama}
\end{table*}

\section{Additional Results}\label{A:extra_experiments}

Additional evaluation is conducted on smaller verifier models \textsc{Qwen-2.5-14B-Instruct} and \textsc{LLaMA-2-13B-chat}. As previously noted, the diffusion drafters used for the 70B+ parameter models are already oversize when comparing to conventional AR draft models. For these smaller verifiers, the 7B parameter diffusion drafters are certainly too large to behave as optimal drafters in this ablation. However, as the optimal diffusion drafter size remains in question, it is valuable insight to see how decoding behavior changes with scale of the verifier.

\Cref{tab:EAGLE2-style-qwen-llama} reports the speed-up and throughput for the smaller verifiers. While \textsc{DiffuCoder} remains a viable drafter for the \textsc{Qwen-2.5-14B-Instruct}, outperforming EAGLE-2 in nearly all settings, the weaker diffusion drafter \textsc{DiffuLlama} is unable to consistently outperform EAGLE-2. With the diffusion draft model being over $20\times$ larger than EAGLE/EAGLE-2 drafters, the drafting latency is no longer competitive, with the higher tokens per draft being the primary factor leading to \textit{SpecDiff-2's} superior performance on select tasks.

\subsection{Llama Model Temperature Ablation}\label{A:extra_experiments-llama}

Supplementing the drafter temperature analysis provided in the main text (\Cref{fig:ssa-scaling}), similar analysis is provided for the \textsc{DiffuLlama} draft model.

\begin{figure}[p]
  \centering
    \includegraphics[height=0.9\panelH]{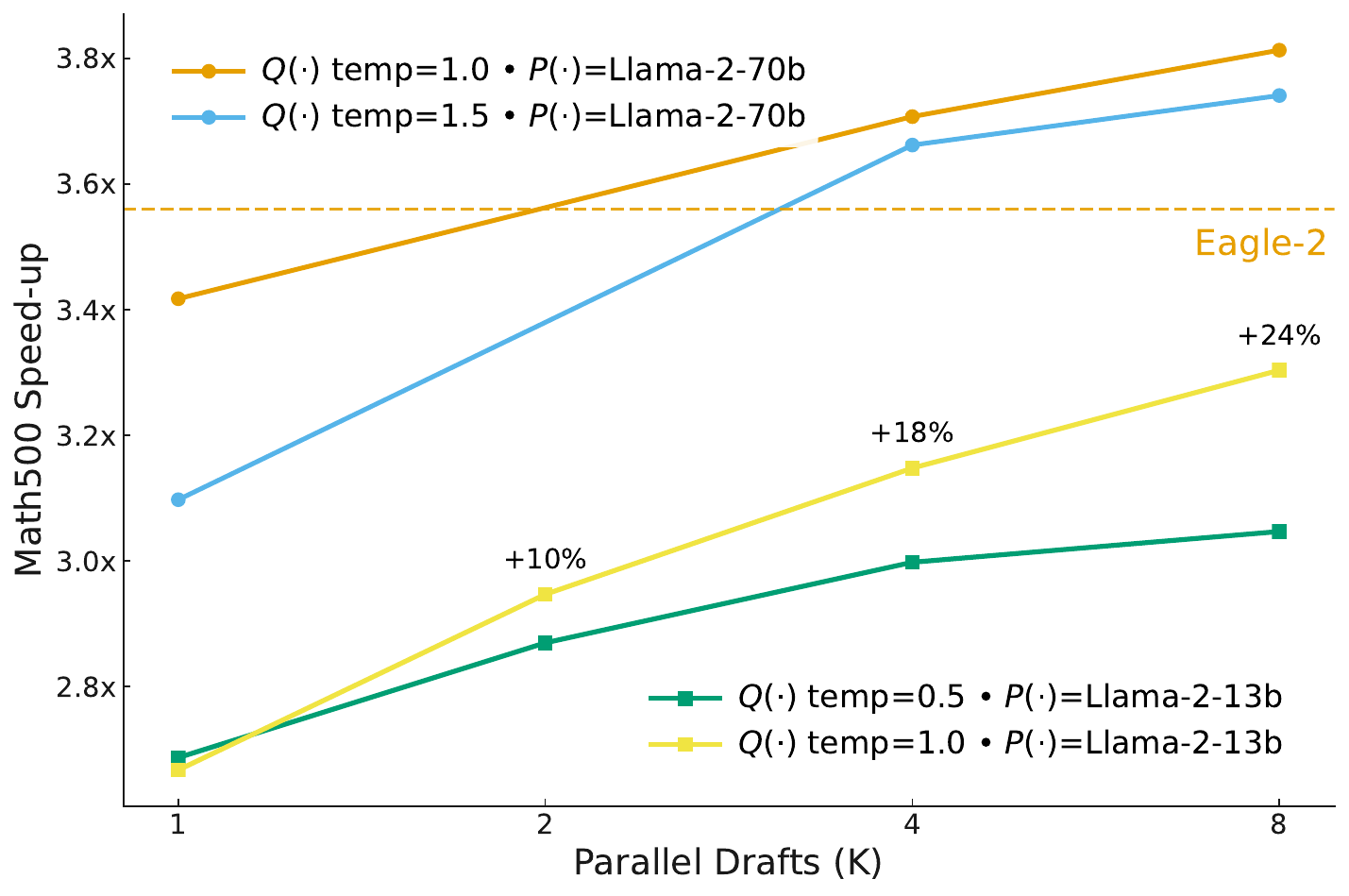}

  \caption{Train-time and test-time alignment methods for all model pairs. We show scaling in speed-up on Math500 prompts as test-time speed-up increases during self-selection acceptance.}
  \label{fig:llama-temp}
\end{figure}

As shown in \Cref{fig:llama-temp}, maintaining $\operatorname{temperature}=1.0$ results in the strongest performance when scaling the number of drafts.

\subsection{Diffusion Steps Ablation}\label{A:extra_experiments-steps}

\begin{figure}[p]
  \centering
    \includegraphics[height=0.9\panelH]{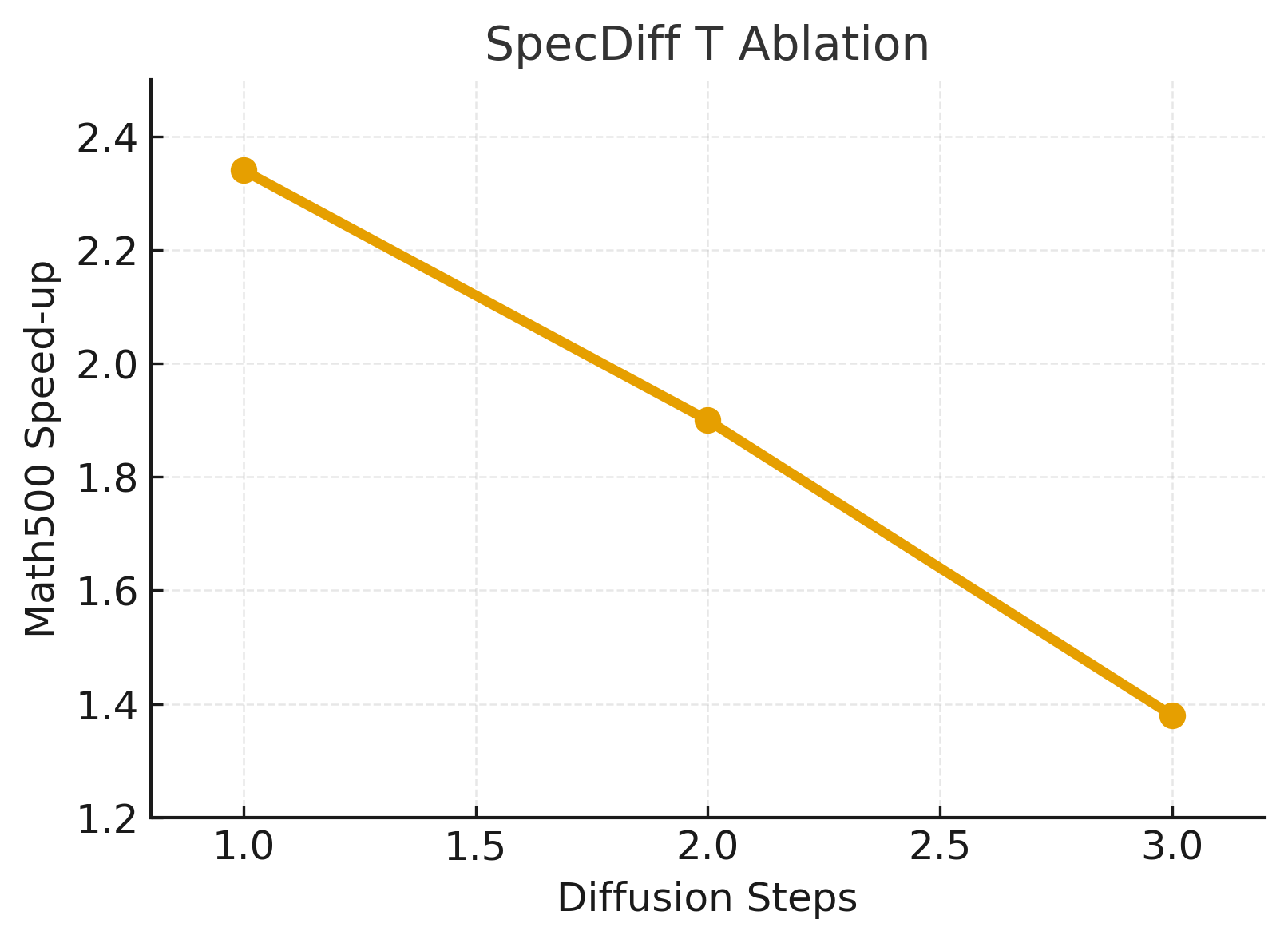}

  \caption{SpecDiff (with an unaligned drafter \textsc{DiffuCoder}) ablated over increasing diffusion steps. Clear downward trend where acceptance vs latency trade-off is undesirable.}
  \label{fig:diffusion-steps}
\end{figure}

As mentioned in the main text, the number of diffusion steps is set to $T=1$. As illustrated by \Cref{fig:diffusion-steps}, speed-up quickly degrades as the number of diffusion steps increases. This is a byproduct of the draft latency increasing linearly with $T$, while the increase in accepted tokens per draft remains relatively stagnant. 

\vspace{-10pt}

\subsection{Gamma Ablation}

\begin{figure}[p]
  \centering
    \includegraphics[height=0.9\panelH]{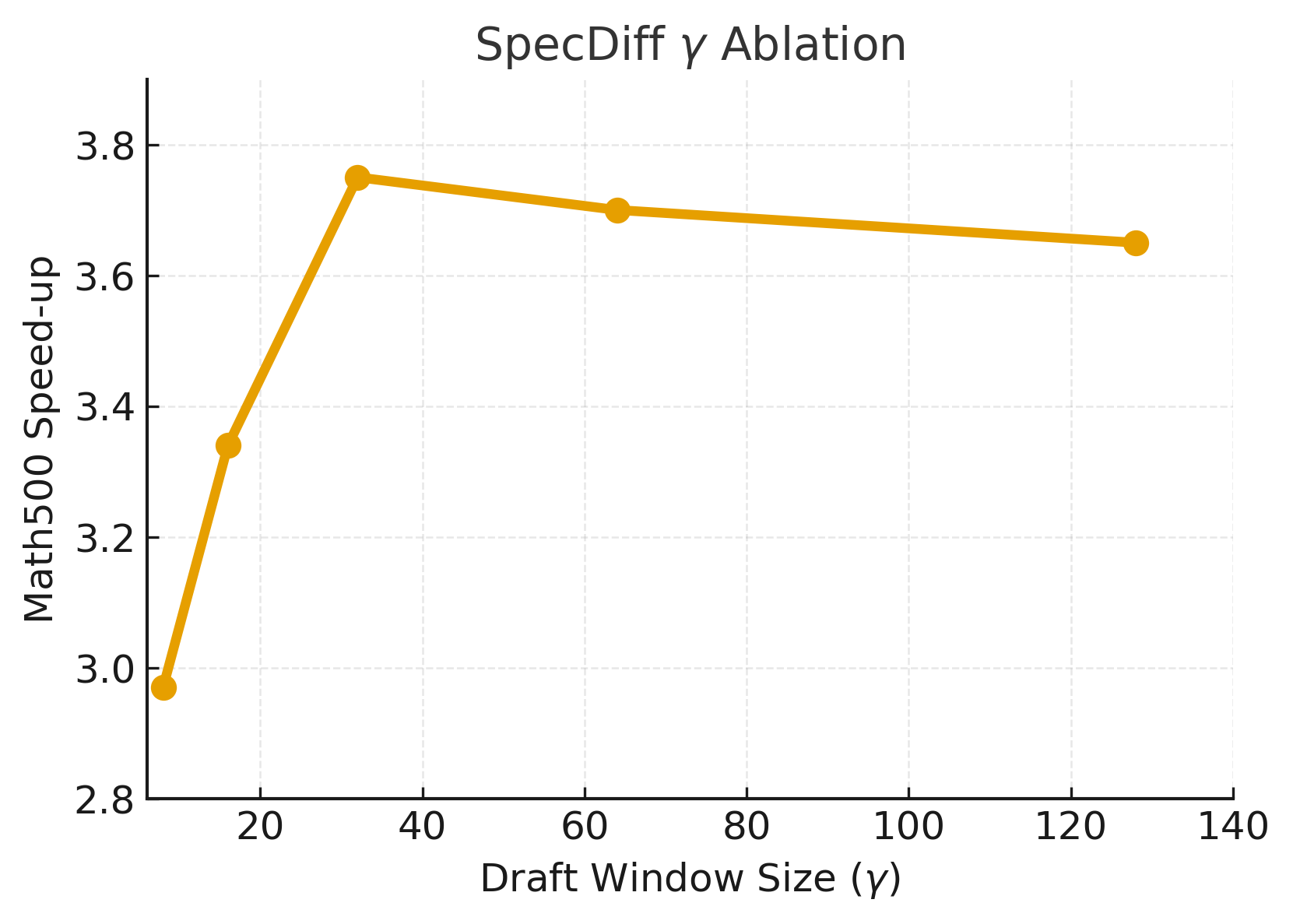}

  \caption{SpecDiff (unaligned drafter - \textsc{DiffuCoder}) ablated over increasing diffusion window size (gamma). Degradation for large gamma, insufficient streak-acceptance for lower gamma.}
  \label{fig:gamma-ablate}
\end{figure}

The experiments use $\gamma=32$ for \textsc{DiffuCoder} and $\gamma=16$ for \textsc{DiffuLLaMA}.
\Cref{fig:gamma-ablate} illustrates how different window sizes influence the overall speed-up for \textsc{DiffuCoder}.
Preliminary experiments revealed that the draft window size $\gamma$ depends largely on the diffusion drafter. 
The window size is set as $\gamma=32$ for \textsc{DiffuCoder} and $\gamma=16$ for \textsc{DiffuLLaMA}. 
These sizes effectively balance between small $\gamma$ values which cannot accept long drafts and large $\gamma$ values which lower draft quality, as ablated in Appendix \ref{A:experimental-details} and established in previous work \citep{christopher2025speculative}. 

Empirical observations suggest \textsc{DiffuCoder} more effectively handles large drafts than \textsc{DiffuLLaMA}, likely due to more extensive pretraining. \textit{\textsc{DiffuCoder} was trained on 130B tokens of specialized data, while DiffuLLaMA was trained on 60B tokens mostly from the web-text corpus SlimPajama.}

\subsection{Joint Alignment Scaling}

\begin{figure}[ht!]
  \centering
  \includegraphics[height=0.8\panelH]{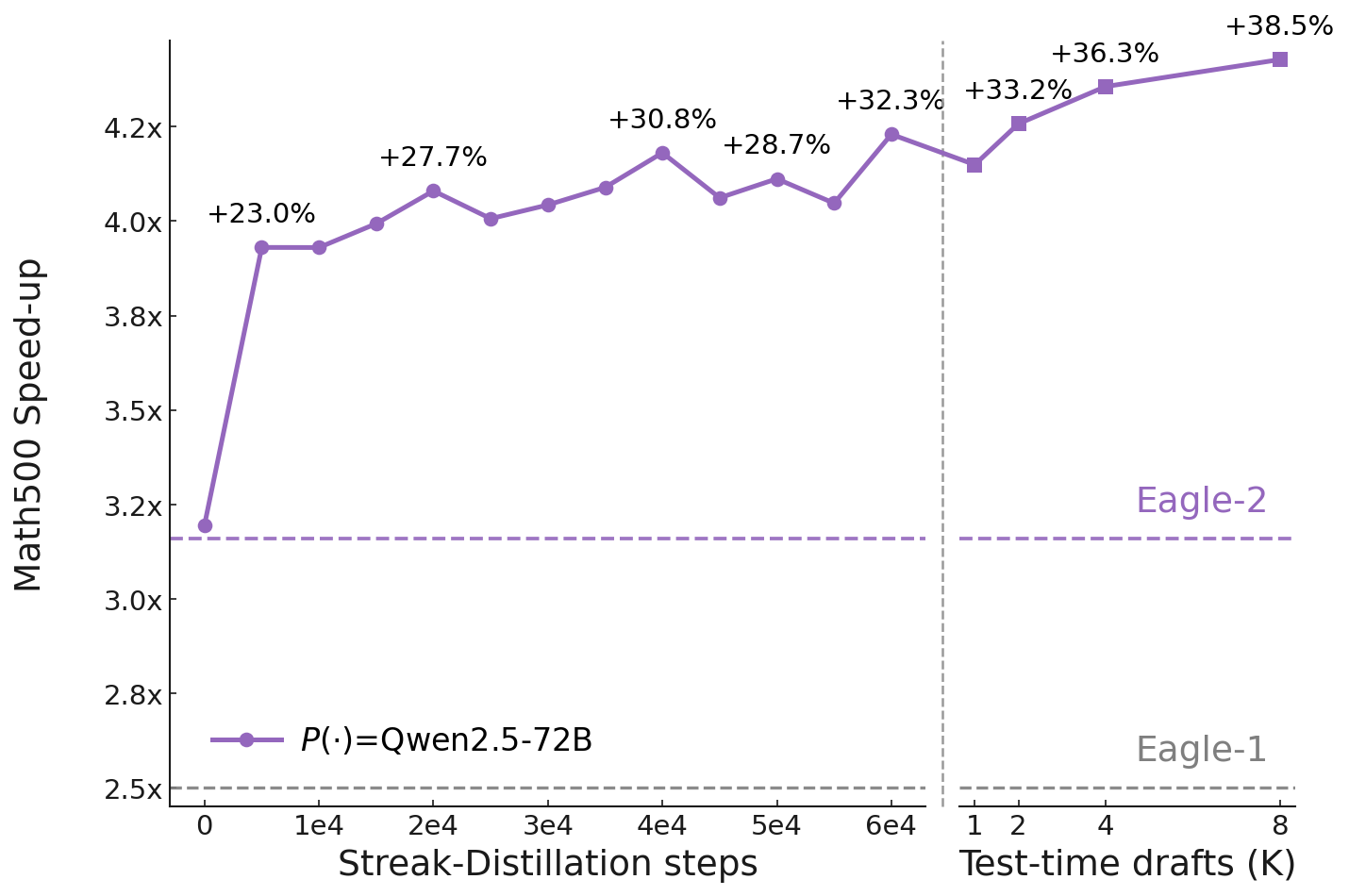}
  \caption{The test-time and train-time phases of \emph{SpecDiff-2.0}. }
  \label{fig:full-alignment-scaling}
\end{figure}

Supplementing the visualizations of train-time and test-time scaling, \Cref{fig:full-alignment-scaling} illustrates the joint impact of applying both alignment techniques.
By combining streak-distillation and parallel drafting for test-time alignment, we see a $+39\%$ improvement over base diffusion drafter, and EAGLE-2 baseline, achieving $~4.3\times$ throughput increase over \textsc{Qwen2.5-72B} model on Math500 data.

\begin{figure*}[t]
  \centering
  \begin{subfigure}[t]{1.0\textwidth}
    \centering
    \includegraphics[height=2.5in]{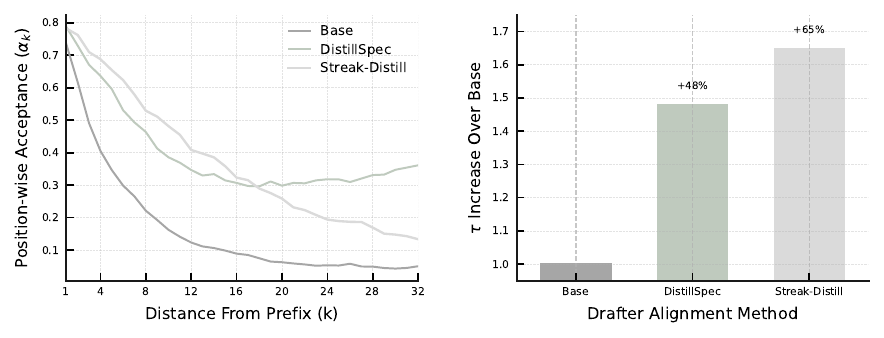}
  \end{subfigure}
  \caption{\textbf{Streak-distillation is compute-effective.} Streak-distillation (Streak-Distill above) compared with DistillSpec at optimizing alignment over the base model. Both algorithms ran for 40k training steps ($\sim$50 Gpu-Hours) on Math500 prompts, for Qwen2.5-14B-Instruct verifier. We see Streak-Distill out-preforms DistillSpec by $+35\%$ alignment/throughput increase ($\tau$ increase above) over the base model, illustrating the weighting issues of naive DistillSpec, and the advantages of the theoretically motivated Streak-Distill.}
  \label{fig:sd_vs_ds} 
  \vspace{10pt}
\end{figure*}

\subsection{Streak-distillation versus DistillSpec}

To further justify the neceesity of streak-distillation, the results in \Cref{fig:sd_vs_ar} are supplemented with another variation of DistillSpec \cite{zhou2023distillspec}.
We refer to the baseline \textit{DistillSpec} in \Cref{fig:sd_vs_ds} as a naive adjustment of the \textit{AR-Distillation} approach in \Cref{fig:sd_vs_ar}.
Specifically, it extends the window to $\gamma$, matching a semi-autoregressive training approach often used for DLMs, where an average over all position-wise acceptances is used for alignment. However, it continues to operate invariant to positional dependencies, given that later tokens are over emphasized relative to their theoretic importance.

The results in \Cref{fig:sd_vs_ds} (left) illustrate that this naive implementation results in significantly lower expected throughput, given mismatch from the theoretic definition of throughput.
This results in the tangible gap between DistillSpec and streak-distillation reported throughputs, shown in \Cref{fig:sd_vs_ds} (right).

\section{Experimental Details}
\label{A:experimental-details}

This appendix documents the hardware, software, model-loading pipeline, and evaluation protocol used in all speculative diffusion experiments. Unless stated otherwise, all numbers reported in the main text were obtained under the configuration described below.

\subsection{Hardware Setup}
All experiments were executed on a single NVIDIA A100 GPU with 80\,GB (two GPUs for 70B+ experiemnts). The GPU was used for inference only; model assembly (including LoRA merging) was performed on CPU first, then the fully assembled drafter model was moved to CUDA \emph{once} to avoid repeated device transfers and to limit allocator fragmentation. CPU nodes were standard x86\_64 Linux servers with at least 256\,GB system memory to accommodate temporary model objects during LoRA merge.

\subsection{Software Environment}
Experiments were run with:
\begin{itemize}
    \item \textbf{Python}: 3.10
    \item \textbf{PyTorch}: 2.3+ (with CUDA enabled)
    \item \textbf{Transformers}: 4.46.x (for both drafter and verifier loading)
    \item \textbf{PEFT}: for loading and merging LoRA adapters
    \item \textbf{Datasets}: Hugging Face \texttt{datasets} for benchmark streaming (MATH-500, LiveCodeBench, GSM8K, MT-Bench)
    \item \textbf{Matplotlib}, \textbf{NumPy}: for lightweight timing/diagnostic plots emitted by the script
\end{itemize}

Mixed-precision and kernel configuration followed:
{\small
\begin{verbatim}
torch.backends.cuda.matmul.allow_tf32 = True
torch.backends.cudnn.allow_tf32 = True
torch.set_float32_matmul_precision("high")
\end{verbatim}}
to enable tensor-core friendly execution on A100 while retaining numerical stability.

\subsection{Models}
\paragraph{Verifier.}
The verifier was loaded with \texttt{AutoModelForCausalLM.from\_pretrained(...)} on GPU, using the same tokenizer and with sampling flags normalized to avoid configuration warnings (e.g.\ forcing \texttt{do\_sample=True} and setting a small but positive temperature). The verifier was evaluated in \texttt{bfloat16} on GPU, with \texttt{max\_memory=\{0: "75GiB", 1: "75GiB"\}} passed to \texttt{from\_pretrained} to make large-model loading explicit on multi-GPU machines.

\paragraph{Drafter.}
The drafter model was loaded through a CPU-first path:
\begin{enumerate}
    \item detect whether \texttt{--model-id} is a \emph{LoRA adapter dir}, a \emph{merged full model}, or a \emph{base model}, by checking for \texttt{adapter\_config.json} and \texttt{config.json};
    \item if an adapter was provided (either as \texttt{--model-id} or as \texttt{--drafter-lora}), attach it to the base model on CPU;
    \item if vocab sizes mismatch, resize the base embeddings to the adapter’s expected vocab and re-attach;
    \item call \texttt{merge\_and\_unload()} so that the final drafter is a single model object;
    \item move the merged model to CUDA once, and cast to the target dtype (typically \texttt{bfloat16}).
\end{enumerate}
This procedure ensures that all speculative runs (and their timing) are not polluted by runtime adapter merges.

\subsection{Acceptance and Distribution Collection}
A key component of the script is the \texttt{\_DraftProbCollector}, which is attached as a hook to the drafter’s generation loop. For each speculative pass, after the drafter emits a block of \(\gamma\) tokens:
\begin{enumerate}
    \item we record the \emph{positions} of newly generated tokens (the ``tail'' of the sequence);
    \item we record the drafter’s per-token probabilities (or log-probabilities) for those positions;
    \item optionally, we record the \emph{full} token distribution for each of the \(\gamma\) positions (not just the chosen token), in model dtype, to enable verifier-side reweighting.
\end{enumerate}
On the verifier side, for each drafted token we compute the verifier distribution, compare it with the drafter distribution, and apply a greedy-style acceptance rule:
\begin{itemize}
    \item if a uniform \(r \sim \mathcal{U}(0,1)\) is less than the verifier probability of the drafted token, the token is \emph{accepted};
    \item otherwise, the token is \emph{replaced} by sampling from a re-normalized verifier distribution in which the rejected draft token is zeroed out.
\end{itemize}
This produces a per-pass \emph{accepted streak} length; over the course of generation we log all streaks and report their mean tokens per draft.

\subsection{Evaluation Protocol}
For each benchmark example \(x\), the script executes:
\begin{enumerate}
    \item \textbf{Speculative run:} call \texttt{speculative\_loop(...)} to generate up to \texttt{--max-new} tokens, logging:
    \begin{itemize}
        \item wall-clock time for drafter forward(s);
        \item wall-clock time for verifier scoring;
        \item KV-cache truncation / advance time;
        \item number of passes and per-pass accepted streaks;
        \item final sequence (for manual auditing).
    \end{itemize}
    \item \textbf{Baseline run (optional):} call \texttt{verifier\_only\_generate(...)} on the \emph{same} prompt with the same sampling parameters; this supplies an actual tokens-per-second baseline for the big model on that hardware.
\end{enumerate}
Throughput is then reported as
\[
\text{throughput} = \frac{\text{new tokens}}{\text{wall time (s)}},
\]
and speculative speed-up is computed as the ratio of baseline throughput to speculative throughput on the same hardware and prompt.

\subsection{Dataset Streaming}
When \texttt{--benchmark} is set, problems are streamed directly from Hugging Face:
\begin{itemize}
    \item \textbf{MATH-500:} \texttt{HuggingFaceH4/MATH-500}, split=\texttt{test}; each problem is wrapped as \texttt{"Problem: <text> \textbackslash n Answer:"}.
    \item \textbf{LiveCodeBench:} \texttt{livecodebench}, \texttt{version\_tag=release\_latest}, including platform and starter code in the prompt.
    \item \textbf{MT-Bench:} \texttt{HuggingFaceH4/mt\_bench\_prompts}; only the first turn is used as a single-turn prompt.
    \item \textbf{GSM8K:} \texttt{openai/gsm8k}, split=\texttt{test}; questions are wrapped as \texttt{"Question: ... \textbackslash n Answer:"}.
\end{itemize}
A global example limit is enforced via \texttt{--limit} to make runs tractable on single GPUs.

\subsection{Datasets}
Distillation datasets were split into three classes. 

\paragraph{(1) Qwen 14B, and 72B corpus.} Corpus of temperature 1.0 completions were generated on prompts from gsm8k, alpaca, and livecodebench. Samples were generated with temp=1.0 by verifiers for smoother distributions, and further generated up-to 32 completions per prompt for distributional coverage, across 128 prompts. Completions were mixed evenly at 33\% per benchmark.  The full corpus of completions were not observed during training, given i.e., distillation compute used was insufficient for a full epoch. Observed high generalization across different tasks. Precise mix appeared un-important.

\paragraph{(2) Llama 13b corpus.} Filtered Llamatoloka/beemo and openbmb/UltraFeedback SFT datasets for filtered 13b completions. Maintained even 50, 50 mix. Specifically, \emph{toloka/beemo} — A curated set of machine-generated responses with per-row model labels; ($\sim$2.19k rows in the default split. \emph{openbmb/UltraFeedback} — Preference dataset with 64k prompts / 256k responses; We filtered to entries generated by LLaMA-2-13B-chat.

\paragraph{(3) Llama 70b corpus.} Completions filtered from togethercomputer/llama-instruct for Llama-2-70b. \emph{togethercomputer/llama-instruct} — 19,004 English instruction–response conversations in [INST]…[/INST] format. 

Generally, we found distillation to be very robust to differing datasets, exhbiting strong transfer.

\subsection{Test-time verification.} 
\label{A:experimental-details-tt}
$K$ distinct drafts may entail redundancy at different token positions. We construct a tree from drafted tokens, pass flattened block to verifier, with custom attention mask. Verifier outputs probability distributions over tokens, conditioned on respective paths. Outputs are cached, and evaluated in parallel. This multi-branch, tree-based test-time verification was implemented using standard PyTorch and Hugging Face tooling, together with a few lightweight utility layers, and was integrated into the same speculative-diffusion pipeline as described. We describe the relevant components below.

\paragraph{Core frameworks.}
All tensor-level operations (tree flattening, parent-index bookkeeping, mask construction, and batched logit gathering) were implemented in \texttt{PyTorch} (2.3+). We relied on PyTorch’s native advanced indexing to map from \emph{tree order} to \emph{flattened order}, and we used batched \texttt{torch.gather} to pull out verifier probabilities corresponding to the drafted tokens at each depth. The verifier itself was loaded via \texttt{transformers.AutoModelForCausalLM} and executed in \texttt{bfloat16} on GPU, so the custom mask had to be materialized on the same device and dtype-compatible with the model’s attention stack (i.e.\ \texttt{bool} or \texttt{int} mask, depending on the architecture).

\paragraph{Tokenizer and model plumbing.}
We kept using \texttt{transformers.AutoTokenizer} to ensure that the flattened sequence \(\tilde{x}\) remained consistent with the underlying causal LM. Since the tree construction introduced non-contiguous branches, we stored an auxiliary Python-side structure:
\begin{verbatim}
List[Dict[str, Tensor]]
\end{verbatim}
for each level of the tree, holding (i) token ids, (ii) branch ids, and (iii) parent indices. This allowed us to re-associate the model’s output logits with the correct logical branch after the forward pass. Because we had already aligned tokenizer and embeddings to avoid vocab mismatches when loading LoRA or merged checkpoints, no extra resizing was needed at this stage.

\paragraph{Mask construction.}
The custom attention mask was built with pure tensor operations. We first created a lower-triangular causal mask of shape \(T \times T\) (where \(T = |\tilde{x}|\)) using \texttt{torch.tril}, then selectively \emph{zeroed out} the entries corresponding to cross-branch attention at the same depth. To do this efficiently, we precomputed, per depth, the flat indices of tokens belonging to each branch and used \texttt{torch.index\_fill\_} (or direct boolean assignment) to disable attention among sibling tokens. On models that expected an additive mask (e.g.\ causal LMs returning \texttt{attn\_weights + mask}), we converted the boolean mask to the large negative form (e.g.\ \(-10^4\) or \(-10^9\)) and broadcast it to the expected number of heads. This was passed to the model via the same keyword arguments the \texttt{transformers} model already supported (typically \texttt{attention\_mask} or \texttt{attention\_bias}, depending on the architecture).

\paragraph{KV-cache reuse.}
We reused the verifier’s KV-cache between speculative passes in exactly the same manner as the single-branch setup: the prefix \(x_{1:L}\) was scored once, and the resulting \texttt{past\_key\_values} object was stored on GPU. For the multi-branch case we only extended the cache along valid branch paths. This was straightforward because the flattened tree preserved topological order; we simply advanced the KV-cache in the same order in which we had flattened the tokens. Where the underlying model exposed a \texttt{.crop()} or similar method (as in our speculative loop), we invoked it to truncate the cache to the accepted depth.

\paragraph{Parallel acceptance and logging.}
After the batched forward, we obtained logits of shape \((T, V)\), where \(V\) is the vocabulary size. We then gathered, for every drafted node, the probability of \emph{its} proposed token and, in parallel, the full verifier distribution for Dirac/SDA variants. This was done with a single \texttt{torch.softmax} call (in \texttt{float32} for numerical safety) followed by masked sampling. Random draws for all nodes at a given depth were produced in batch using \texttt{torch.rand} on GPU, so accept/reject decisions could be applied with simple elementwise comparisons. All acceptance decisions, together with branch ids and depths, were written to the same JSONL logging utility we already used for the speculative loop, so later analysis could reconstruct which branches were accepted and at which depth.

\paragraph{Profiling utilities.}
To ensure that the tree flattening and mask construction did not dominate runtime, we added lightweight timing scopes (via \texttt{time.perf\_counter()}) around:
\begin{enumerate}
    \item tree build,
    \item mask materialization,
    \item verifier forward.
\end{enumerate}
These were recorded in the per-run \texttt{timers} dictionary alongside the existing entries for “drafter,” “verifier\_score,” and “kv\_truncate,” so that the overhead of multi-branch verification could be compared 1:1 with the single-branch speculative loop reported in the main experiments.

\paragraph{Optional optimizations.}
On runs where the flattened sequence became long (large \(K\), large \(\gamma\)), we exploited:
\begin{itemize}
    \item \textbf{in-place mask reuse}: for fixed \(K\) and \(\gamma\), the mask pattern per depth was constant, so we cached it and only re-filled the branch-specific indices;
    \item \textbf{AMP / autocast}: where the model permitted it, we wrapped the verifier forward in \texttt{torch.autocast("cuda", dtype=torch.bfloat16)} to reduce memory pressure;
    \item \textbf{SDPA backends}: we kept the SDPA context from the original script to allow Flash / memory-efficient attention, controlled by the \texttt{--no-flash} flag.
\end{itemize}

Overall, this combination of (i) PyTorch tensor operations, (ii) Hugging Face causal LMs, (iii) explicit attention-mask construction, and (iv) the existing JSONL logging infrastructure made the multi-draft verification slot cleanly into the rest of the speculative decoding framework, while still letting us profile and report speed-ups in a reproducible manner.

\section{Complete Proofs}

\begin{theorem}[SD-Invariance]\label{A:sd-invariance}  

Tokens sampled via \emph{speculative sampling} from $p(x)$ and $q(x)$ are distributed identically to those sampled from $p(x)$. 

\begin{proof}
(As reported in \citep{leviathan2023fast}) Let $\alpha$ be the acceptance probability. Note that as $p'(x) = \operatorname{norm}(\max(0, p(x) - q(x))) = \frac{p(x) - \min(q(x), p(x))}{\sum_{x'}(p(x') - \min(q(x'), p(x')))} = \frac{p(x) - \min(q(x), p(x))}{1 - \alpha}$, the normalizing constant for the adjusted distribution $p'(x)$ is $1 - \alpha$.

Now:

\begin{align*}
    P(x=x') = P(\text{guess}\ \text{accepted}, x=x') +\\ P(\text{guess}\ \text{rejected}, x=x')
\end{align*}
Where:
\begin{align}
    P(\text{guess}&\ \text{accepted}, x=x') = q(x')\min(1, \frac{p(x')}{q(x')})\\ &\;\;\;\;\;\;\;\;\;\;\;\;\;\;\;\;\;\;\;\;\;\;\;\;= \min(q(x'), p(x'))\\
    P(\text{guess}&\ \text{rejected}, x=x') = (1 - \alpha)p'(x')\\  
    &\;\;\;\;\;\;\;\;\;= p(x') - \min(q(x'), p(x'))
\end{align}
And thus, overall we obtained the sought result:
\begin{align*}
    P(x=x') = \min(p(x'), q(x')) + p(x')&\\
    - \min(p(x'), q(x')) = p(x').
\end{align*}
\end{proof}
\end{theorem}

\begin{theorem}[Position-Wise Alpha Identity]\label{A:pos_alpha_identity}
Let position wise acceptance rates $\alpha_1, \ldots, \alpha_\gamma$ be exchangable (i.e., as is asumed for AR drafters), then maximizing $\frac{\text{Tokens}}{\text{Draft}}$ reduces to $\max[\alpha_1({\bm{s}} \circ \bm{x}_{1})]$ (acceptance directly proceeding prefix $\bm{s}$.
\begin{proof}
     \begin{align*}
    \max[\sum_{m=1}^{\gamma}\prod_{j=1}^{m} \alpha_j({\bm{s}} \circ \bm{x}_{1:j})]&=\max[\sum_{m=1}^{\gamma}\prod_{j=1}^{m} \alpha_j({\bm{s}} \circ \bm{x}_{1:j})],\\
    &=\max[\sum_{m=1}^{\gamma}\alpha_j({\bm{s}} \circ \bm{x}_{1:j})^m],\\
    &=\sum_{m=1}^{\gamma}\max[\alpha_j({\bm{s}} \circ \bm{x}_{1:j})]^m,\\
    &\Rightarrow\sum_{m=1}^{\gamma}\max[\alpha_j({\bm{s}} \circ \bm{x}_{1:j})],\\
    &\Rightarrow \max[\alpha_1({\bm{s}} \circ \bm{x}_{1})].
    \end{align*}
\end{proof}
\end{theorem}

\section{Model Completions}\label{A:example_completions}
\subsection{Qwen2.5-72B-Instruct}
Sampled at temperature 0.0, at a maximum of 3096 tokens. <thinking> tokens omitted, reasoning formatted below, answers instructed to be in `boxed\{\}' format. Prompts sampled from math500 corpus.

\subsection*{Problem 1: Algebra}

\textbf{Problem.} Hillary has eleven coins, all dimes and nickels. In total, the coins are worth 75 cents. How many nickels does she have?

\textbf{Solution.}
Let \( d \) represent the number of dimes and \( n \) represent the number of nickels.
\[
\begin{aligned}
d + n &= 11,\\
10d + 5n &= 75.
\end{aligned}
\]
Dividing the second equation by 5 gives \(2d + n = 15\).  
Subtracting the first equation from this, we obtain \(d = 4\).  
Substitute back: \(4 + n = 11 \implies n = 7.\)

\textbf{Answer.} \(\boxed{7}\)

\subsection*{Problem 2: Intermediate Algebra}

\textbf{Problem.} Find the maximum value of
\[
f(x,y) = x\sqrt{1 - y^2} + y\sqrt{1 - x^2}, \quad -1 \le x,y \le 1.
\]

\textbf{Solution.}
Let \(x = \cos\theta, \, y = \cos\phi\). Then \(\sqrt{1-y^2}=\sin\phi\) and \(\sqrt{1-x^2}=\sin\theta\):
\[
f = \cos\theta\sin\phi + \cos\phi\sin\theta = \sin(\theta+\phi).
\]
The maximum of \(\sin(\theta+\phi)\) is \(1\), which is attainable (e.g., \(x=1,\,y=0\)).

\textbf{Answer.} \(\boxed{1}\)

\subsection*{Problem 3: Algebra}

\textbf{Problem.} For how many values of \(x\) is the expression
\[
\frac{x-5}{x^2-9}
\]
not defined?

\textbf{Solution.}
The denominator is zero when \(x^2 - 9 = 0 \Rightarrow (x-3)(x+3)=0\).  
Thus \(x = 3\) or \(x = -3\).

\textbf{Answer.} \(2\)

\subsection*{Problem 4: Geometry}

\textbf{Problem.} Altitudes \(\overline{AD}\) and \(\overline{BE}\) of \(\triangle ABC\) intersect at \(H\).  
If \(\angle BAC = 54^\circ\) and \(\angle ABC = 52^\circ\), find \(\angle AHB\).

\textbf{Solution.}
\[
\angle ACB = 180^\circ - 54^\circ - 52^\circ = 74^\circ.
\]
In \(\triangle ABC\), \(\angle AHB = 180^\circ - \angle ACB = 106^\circ.\)

\textbf{Answer.} \(\boxed{106^\circ}\)

\subsection*{Problem 5: Precalculus}

\textbf{Problem.} Two parallel lines are defined by
\[
\begin{pmatrix}3 \\ -10 \\ 1\end{pmatrix} + t\begin{pmatrix}2 \\ -9 \\ -2\end{pmatrix}, \quad
\begin{pmatrix}-5 \\ -3 \\ 6\end{pmatrix} + u\begin{pmatrix}4 \\ -18 \\ -4\end{pmatrix}.
\]
Find the distance between the lines.

\textbf{Solution.}
Direction vectors are proportional: \(\mathbf{b}_2 = 2\mathbf{b}_1\), confirming parallelism.  
For \(\mathbf{a}_1=(3,-10,1)\), \(\mathbf{a}_2=(-5,-3,6)\), and \(\mathbf{b}_1=(2,-9,-2)\),
\[
d = \frac{\|(\mathbf{a}_2-\mathbf{a}_1)\times \mathbf{b}_1\|}{\|\mathbf{b}_1\|}.
\]
Compute:
\[
(\mathbf{a}_2-\mathbf{a}_1)\times \mathbf{b}_1 = (31,-6,58), \quad
\|\mathbf{b}_1\|=\sqrt{89}.
\]
Hence
\[
d = \frac{\sqrt{4361}}{\sqrt{89}} = \sqrt{49} = 7.
\]

\textbf{Answer.} \(\boxed{7}\)

\subsection*{Problem 6: Precalculus}

\textbf{Problem.} Two runners \(A\) and \(B\) start at point \(O\) and run in the same direction.  
Runner \(B\) runs three times as fast as \(A\).  
An observer at point \(P\) (with \(\overline{OP}\) perpendicular to the track) observes angle \(\angle APB\).  
Find the maximum of \(\angle APB\) (in degrees).

\textbf{Solution.}
At time \(t\), \(A(vt,0)\), \(B(3vt,0)\), \(P(0,d)\).  
The angle satisfies
\[
\theta = \arctan\!\left(\frac{d}{vt}\right) - \arctan\!\left(\frac{d}{3vt}\right).
\]
Maximizing over \(t>0\) yields \(\theta_{\max} = 30^\circ.\)

\textbf{Answer.} \(\boxed{30^\circ}\)

\subsection*{Problem 7: Counting and Probability}

\textbf{Problem.} In how many ways can 5 students be selected from a group of 6?

\textbf{Solution.}
\[
\binom{6}{5} = \frac{6!}{5!(1!)} = 6.
\]

\textbf{Answer.} \(\boxed{6}\)

\subsection*{Problem 8: Counting and Probability}

\textbf{Problem.} A prodigy wishes to arrange 2 indistinguishable IMO and 2 indistinguishable IPhO gold medals in a row.  
How many distinct arrangements are possible?

\textbf{Solution.}
\[
\text{Distinct permutations} = \frac{4!}{2!2!} = 6.
\]

\textbf{Answer.} \(\boxed{6}\)

\subsection*{Problem 9: Number Theory}

\textbf{Problem.} If Rosa’s age leaves remainder 1 when divided by 2, 3, 4, or 6, and remainder 0 when divided by 7,  
and she is younger than 75, how old is she?

\textbf{Solution.}
From the first set: \(R\equiv1\pmod{12}\).  
Also \(R\equiv0\pmod{7}\).  
Solving \(12k+1\equiv0\pmod7\Rightarrow k\equiv4\pmod7\).  
Thus \(R=12(7m+4)+1=84m+49<75\Rightarrow m=0\).  
Hence \(R=49.\)

\textbf{Answer.} \(\boxed{49}\)

\end{document}